\documentclass[12pt,oneside]{report}

\newcommand{\reporttitle}{Vision-Based Autonomous Drone Control using Supervised Learning in Simulation}
\newcommand{\reportauthor}{Max Christl}
\newcommand{\supervisor}{Professor Wayne Luk}
\newcommand{\degreetype}{MSc Computing Science}

\usepackage[a4paper,hmargin=2.8cm,vmargin=2.0cm,includeheadfoot]{geometry}
\usepackage{textpos}
\usepackage[numbers]{natbib} 
\usepackage{tabularx,longtable,multirow,subfigure,caption}
\usepackage{fncylab} 
\usepackage{fancyhdr} 
\usepackage{url} 
\usepackage[english]{babel}
\usepackage{amsmath}
\usepackage{graphicx}
\usepackage{dsfont}
\usepackage{backref} 
\usepackage{array}
\usepackage{latexsym}

\def\@makechapterhead#1{%
  \vspace*{10\p@}%
  {\parindent \z@ \raggedright \sffamily
    \interlinepenalty\@M
    \Huge\bfseries \thechapter \space\space #1\par\nobreak
    \vskip 30\p@
  }}

\def\@makeschapterhead#1{%
  \vspace*{10\p@}%
  {\parindent \z@ \raggedright
    \sffamily
    \interlinepenalty\@M
    \Huge \bfseries  #1\par\nobreak
    \vskip 30\p@
  }}

\allowdisplaybreaks

\usepackage{verbatim} 
\usepackage[toc,page]{appendix} 
\usepackage{listings} 
\lstdefinestyle{xml}{ 
    basicstyle=\scriptsize,
}
\lstdefinestyle{python}{ 
    basicstyle=\scriptsize,
}
\usepackage{tikz} 
\usepackage{tikz-uml} 
\tikzumlset{font=\tiny} 
\usepackage{forest} 
\usetikzlibrary{arrows.meta,shadows.blur}
\usepackage{pgfplots} 
\usepackage{csquotes} 
\usepackage{algorithm} 
\usepackage{algpseudocode} 

\algnewcommand\algorithmicforeach{\textbf{for each}}
\algdef{S}[FOR]{ForEach}[1]{\algorithmicforeach\ #1\ \algorithmicdo}
\newcommand{\Break}{\State \textbf{break} }
\let\oldReturn\Return
\renewcommand{\Return}{\State\oldReturn}

\usepackage{textcomp} 
\usepackage{gensymb} 
\usepackage{svg} 
\usepackage{footnote} 
\newenvironment{conditions}
  {\par\vspace{\abovedisplayskip}\noindent\begin{tabular}{>{$}l<{$} @{${}={}$} l}}
  {\end{tabular}\par\vspace{\belowdisplayskip}}
 \usepackage{threeparttable} 
\usepackage{booktabs, caption, makecell}
\usepackage[T1]{fontenc}
\usepackage{setspace}

\date{4. September 2020}

\begin{document}

\begin{titlepage}

\newcommand{\HRule}{\rule{\linewidth}{0.5mm}} 


\includegraphics[width = 4cm]{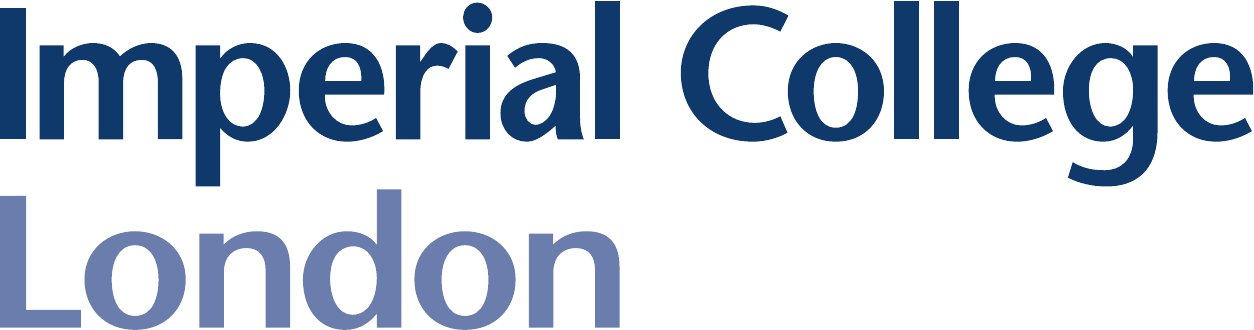}\\[0.5cm] 

\center 


\textsc{\Large Imperial College London}\\[0.5cm] 
\textsc{\large Department of Computing}\\[0.5cm] 


\HRule \\[0.4cm]
{ \huge \bfseries \reporttitle}\\ 
\HRule \\[1.5cm]
 

\begin{minipage}{0.4\textwidth}
\begin{flushleft} \large
\emph{Author:}\\
\reportauthor 
\end{flushleft}
\end{minipage}
~
\begin{minipage}{0.4\textwidth}
\begin{flushright} \large
\emph{Supervisor:} \\
\supervisor 
\end{flushright}
\end{minipage}\\[4cm]

\vfill 
Submitted in partial fulfillment of the requirements for the MSc degree in
\degreetype~of Imperial College London\\[0.5cm]

\makeatletter
\@date 
\makeatother

\end{titlepage}

\pagenumbering{roman}
\clearpage{\pagestyle{empty}\cleardoublepage}
\setcounter{page}{1}
\pagestyle{fancy}

\begin{abstract}
Limited power and computational resources, absence of high-end sensor equipment and GPS-denied environments are challenges faced by autonomous micro areal vehicles (MAVs). We address these challenges in the context of autonomous navigation and landing of MAVs in indoor environments and propose a vision-based control approach using Supervised Learning. To achieve this, we collected data samples in a simulation environment which were labelled according to the optimal control command determined by a path planning algorithm. Based on these data samples, we trained a Convolutional Neural Network (CNN) that maps low resolution image and sensor input to high-level control commands. We have observed promising results in both obstructed and non-obstructed simulation environments, showing that our model is capable of successfully navigating a MAV towards a landing platform. Our approach requires shorter training times than similar Reinforcement Learning approaches and can potentially overcome the limitations of manual data collection faced by comparable Supervised Learning approaches.
\end{abstract}

\cleardoublepage
\section*{Acknowledgements}
I would like to express my sincere gratitude to Professor Wayne Luk and Dr. Ce Guo for their thought-provoking input and continuous support throughout the project.

\clearpage{\pagestyle{empty}\cleardoublepage}

\fancyhead[RE,LO]{\sffamily {Table of Contents}}
\tableofcontents 

\listoffigures

\listoftables

\clearpage{\pagestyle{empty}\cleardoublepage}
\pagenumbering{arabic}
\setcounter{page}{1}
\fancyhead[LE,RO]{\slshape \rightmark}
\fancyhead[LO,RE]{\slshape \leftmark}


\chapter{Introduction}

The autonomous control of unmanned areal vehicles (UAVs) has gained increasing importance due to the rise of consumer, military and smart delivery drones. 

Navigation for large UAVs which have access to the Global Positioning system (GPS) and a number of high-end sensors is already well understood \citep{zhang, cho, sukkarieh, hoffmann, langel}. These navigation systems are, however, not applicable for MAVs which only have limited power and computing resources, aren't equipped with high-end sensors and are frequently required to operate in GPS-denied indoor environments. A solution to the autonomous control of MAVs are vision-based systems that only rely on data captured by a camera and basic onboard sensors. 

An increasing number of publications have applied machine learning techniques for vision-based MAV navigation. These can generally be differentiated into Supervised Learning approaches that are trained based on labelled images captured in the real world \citep{kim, ram} and Reinforcement Learning approaches which train a machine learning model in simulation and subsequently transfer the learnings to the real world using domain-randomisation techniques \citep{sadeghi, kaufmann, polvara}. The former approach is limited by the fact that the manual collection and labelling of images in the real world is a time consuming task and the resulting datasets used for training are thus relatively small. The latter approach shows very promising results both in simulation and in the real world. A potential downside is, however, the slow conversion rate of Reinforcement Learning algorithms that requires the models to be trained for several days. 

We are suggesting an approach that combines the strengths of Supervised Learning (i.e. a quick convergence rate) with the advantages of training a model in simulation (i.e. automated data collection). The task which we attempt to solve is to autonomously navigate a drone towards a visually marked landing platform. Our approach works as follows: We capture images (and sensor information) in simulation and label them according to the optimal flight command that should be executed next. The optimal flight command is computed using a path planning algorithm that determines the optimal flight path towards the landing platform. The data collection procedure is fully automated and thus allows for the creation of very large datasets. These datasets can subsequently be used to train a CNN through Supervised Learning which significantly reduces the training time. These learnings can then be transferred into the real world under the assumption that domain randomisation techniques are able to generalize the model such that it is able to perform in unseen environments. We have refrained from applying domain randomisation techniques and only tested our approach in simulation due to restrictions imposed by the COVID-19 pandemic. Previous research has shown that sim-to-real transfer is possible using domain randomisation \citep{sadeghi, kaufmann, polvara, tobin, johns}.

This report makes two principal contributions. First, we provide a simulation program which can be used for simulating the control commands of the DJI Tello drone in an indoor environment. This simulation program is novel as it is specifically targeted at the control commands of the DJI Tello drone, provides the possibility of capturing images from the perspective of the drone's camera, can be used for generating labelled datasets and can simulate flights that are controlled by a machine learning model. Second, we propose a vision-based approach for the autonomous navigation of MAVs using input form a low resolution monocular camera and basic onboard sensors. A novelty of our approach is the fact that we collect labelled data samples in simulation using our own data collection procedure (Section~\ref{sec:dataset}). These data samples can subsequently be used for training a CNN through Supervised Learning. Our approach thus manages to overcome the limitations of manual data collection in the real world and the slow convergence rate of simulation-based Reinforcement Learning in regards to the autonomous navigation of MAVs. 

The report is structured in the following way: After giving an overview of previous research in this area, we introduce the simulation program in Chapter~\ref{chr:simulation} and describe our vision-based control approach in Chapter~\ref{chr:ml}. Chapter~\ref{chr:evaluation} gives an overview of the test results that have been achieved by the model in our simulation environment as well as a qualitative comparison with previous publications. The final chapter of this report sets out the conclusion. 
\chapter{Background and Related Literature}

This chapter gives an overview of background literature and related research that is relevant to this project. The first section introduces literature referring to autonomous control of UAVs. And the second part focuses on literature referring to path planning techniques. 

\section{Autonomous UAV Control}

Autonomous control of UAVs is an active area of research. Several techniques for navigating and landing UAVs exist, which can most generally be divided into external positioning system based approaches, infrastructure-based approaches and vision-based approaches. 

External positioning systems such as the Global Positioning System (GPS) can either be used on their own \citep{zhang, cho}, or in combination with onboard sensors such as inertial measurement units (IMUs) \citep{sukkarieh, hoffmann} or internal navigation systems (INSs) \citep{langel} for the autonomous control of UAVs. These navigation approaches are, however, limited by the fact that GPS signals aren't always available, especially in indoor environments. 

A further technique for UAV control, is to use external sensors (i.e. infrared lamps) placed in the infrastructure surrounding the UAV \citep{gui}. This technique allows to accurately estimate the pose of the UAV, but is limited to its purpose build environment. 

Vision-based navigation exclusively uses onboard sensors and as its name suggests in particular visual information. This technique thus overcomes both of the previously mentioned limitations as it can be applied in GPS-denied environments and doesn't rely on specific infrastructures. A widely used vision-based navigation approach is visual simultaneous localization and mapping (visual SLAM) \citep{lu}. \citet{celik} have developed a visual SLAM system for autonomous indoor environment navigation using a monocular camera in combination with an ultrasound altimeter. Their system is capable of mapping obstacles in its surrounding via feature points and straight architectural lines \citep{celik}. \citet{grzonka} introduced a multi-stage navigation solution combining SLAM with path planning, height estimation and control algorithms. A limitation of visual SLAM is the fact that it is computationally expensive and thus requires the UAV to have access to sufficient power sources and high-performance processors \citep{grzonka, aulinas}.  

In recent years, an ever increasing number of research papers have emerged which use machine learning systems for vision-based UAV navigation. \citet{ross} developed a system based on the DAgger algorithm \citep{gordon}, an iterative training procedure, that can navigate a UAV through trees in a forest. \citet{sadeghi} proposed a learning method called CAD$^2$RL for collision avoidance in indoor environments. Using this method, they trained a CNN exclusively in a domain randomised simulation environment using Reinforcement Learning and subsequently transferred the learnings into the real world \citep{sadeghi}. \citet{kaufmann} further extended this idea and developed a modular system for autonomous drone racing that combines model-based control with CNN-based perception awareness. They trained their system in both static and dynamic simulation environments using Imitation Learning and transferred the learnings into the real world through domain randomisation techniques \citep{kaufmann}. \citet{polvara} proposed a modular design for navigating a UAV towards a marker that is also trained in a domain randomised simulation environment. Their design combines two Deep-Q Networks (DQNs) with the first one being used to align the UAV with the marker and the second one being used for vertical descent \citep{polvara}. The task which \citet{polvara} attempt to solve is similar to ours, but instead of a Reinforcement Learning approach, we are using a Supervised Learning approach. 

There also exists research in which Supervised Learning has been applied for vision-based autonomous UAV navigation. \citet{kim} proposed a system made up of a CNN that has been trained using a labelled dataset of real world images. The images were manually collected in seven different locations and labelled with the corresponding flight command executed by the pilot \citep{kim}. A similar strategy for solving the task of flying through a straight corridor without colliding with the walls has been suggested by \citet{ram}. The authors also manually collected the images in the real world by taking multiple images at different camera angles and used them to train a CNN which outputs the desired control command \citep{ram}. Our approach differs regarding the fact that we are collecting the labelled data samples in a fully automated way in simulation. This enables us to overcome the tedious task of manually collecting and labelling images, allowing us to build significantly larger datasets. Furthermore, our environment includes obstacles other than walls that need to be avoided by the UAV in order to reach a visually marked landing platform. 

\section{Path Planning}

Path planning is the task of finding a path from a given starting point to a given end point which is optimal in respect to a given set of performance indicators \citep{lu}. It is very applicable to UAV navigation for example when determining how to efficiently approach a landing area while at the same time avoiding collision with obstacles. \citet{lu} divide literature relating to path planning into two domains, global and local path planning, depending on the amount and information utilised during the planning process. Local path planning algorithms have only partial information about the environment and can react to sudden changes, making them especially suitable for real-time decision making in dynamic environments \citep{lu}. Global path planning algorithms have in contrast knowledge about the entire environment such as the position of the starting point, the position of the target point and the position of all obstacles \citep{lu}. For this project, only global path planning is relevant since we utilise it for labelling data samples during our data collection procedure and have a priori knowledge about the simulation environment.

In regards to global path planning, there exist heuristic algorithms which try to find the optimal path through iterative improvement, and intelligent algorithms which use more experimental methods for finding the optimal path \citep{lu}. We will limit our discussion to heuristic algorithms since these are more relevant to this project. A common approach is to model the environment using a graph consisting of nodes and edges. Based on this graph, several techniques can be applied for finding the optimal path. The most simple heuristic search algorithm is breadth-first search (BFS) \citep{cormen}. This algorithm traverses the graph from a given starting node in a breadth-first fashion, thus visiting all nodes that are one edge away before going to nodes that are two edges away \citep{cormen}. This procedure is continued until the target node is reached, returning the shortest path between these two nodes \citep{cormen}. The Dijkstra algorithm follows a similar strategy, but it also takes positive edge weights into consideration for finding the shortest path \citep{musliman}. The A* algorithm further optimised the search for the shortest path by not only considering the edge weights between the neighbouring nodes but also the distance towards the target node for the decision of which paths to traverse first \citep{hart, filippis}. The aforementioned  algorithms are the basis for many more advanced heuristic algorithms \citep{lu}. 

\citet{szczerba} expanded the idea of the A* algorithm by proposing a sparse A* search algorithm (SAS) which reduces the computational complexity by accounting for more constraints such as minimum route leg length, maximum turning angle, route distance and a fixed approach vector to the goal position. \citet{koenig} suggested a grid-based path planning algorithm called D* Lite which incorporates ideas from the A* and D* algorithms, and subdivides 2D space into grids with edges that are either traversable or untraversable. \citet{carsten} expanded this idea into 3D space by proposing the 3D Field D* algorithm. This algorithm uses interpolation to find an optimal path which doesn't necessarily need to follow the edges of the grid \citep{carsten}. \citet{xiao} also approximate 3D space using a grid-based approach which subdivides the environment into voxels. After creating a graph representing the connectivity between free voxels using a probabilistic roadmap method, \citet{xiao} apply the A* algorithm to find an optimal path. Our path planning approach which will be described in Section~\ref{sec:bfs} is similar to the two previously mentioned algorithms and other grid-based algorithms regarding the fact that we also subdivide space into a 3D grid. But instead of differentiating between traversable and untraversable cells which are subsequently connected in some way to find the optimal path, we use the cells of the grid to approximate the poses that can be reached through the execution of a limited set of flight commands. 

\section{Ethical Considerations}

This is a purely software based project which doesn't involve human participants or animals. We did not collect or use personal data in any form. 

The majority of our experiments were performed in a simulated environment. For the small number of experiments that were conducted on a real drone, we ensured that all safety and environmental protection measures were met by restricting the area, adding protectors to the rotors of the drone, removing fragile objects and other potential hazards as well as implementing a kill switch that forces the drone to land immediately.

Since our project is concerned with the autonomous control of MAVs, there is a potential for military use. The nature of our system is passive and non-aggressive with the explicit objective to not cause harm to humans, nature or any other forms of life and we ensured to comply with the ethically aligned design stated in the comprehensive AI guidelines \citep{ethics}.

This project uses open-source software and libraries that are licensed under the BSD-2 clause \citep{bsd}. Use of this software is admitted for academic purposes and we made sure to credit the developers appropriately. 
\chapter{Simulation}
\label{chr:simulation}

This chapter gives an overview of the drone simulator. The simulator has been specifically developed for this project and is capable of simulating the control commands and subsequent flight and image capturing behaviour of the DJI Tello drone. In the following sections, after laying out the purpose of the simulator and the challenges it needs to overcome, the implementation details of the simulation program are described in regards to the architecture, the drone model, the simulation of movement and the simulation environment.

\section{Challenges}

The reason for building a drone simulator is to automate and speed up the data collection procedure. Collecting data in the real world is a tedious task which has several limitation. These limitations include battery constraints, damages to the physical drone, irregularities introduced by humans in the loop and high cost for location rental as well as human labour and technology expenses. All these factors make it infeasible to collect training data in the real world. In contrast to this, collecting data samples in a simulated environment is cheap, quick and precise. It is cheap since the cost factors are limited to computational costs, it is quick since everything is automated and it is precise because the exact position of every simulated object is given in the simulation environment. In addition, the model can be generalized such that it is capable of performing in unseen environments and uncertain conditions through domain randomisation. This helps to overcome the assumptions and simplifications about the real world which are usually introduced by simulators. For this project, we have however restrained from introducing domain randomisation as we weren't able to transfer the learnings into the real world due to restrictions imposed by COVID-19. Without the possibility of validating a domain randomisation approach, we agreed to focus on a non-domain randomised simulation environment instead. 

Because of all the above mentioned advantages, data collection was performed in simulation for this project. Commercial and open source simulators already exist, such as Microsoft SimAir \citep{airsim} or DJI Flight Simulator \citep{flightsim}. These are, however, unsuitable for this project as they come with a large overhead and aren't specifically targeted at the control commands of the DJI Tello drone. It was therefore necessary to develop a project specific drone simulator which overcomes the following challenges:

\begin{enumerate}

\item{The simulator needs to be specifically targeted at the DJI Tello drone. In other words, the simulator must include a drone with correct dimensions, the ability of capturing image/sensor information and the ability to execute flight commands which simulate the movement of the DJI Tello flight controller.}

\item{The simulator needs to include a data collection procedure which can automatically create a labelled dataset that can be used for training a machine learning model via Supervised Learning.}

\item{The simulator should be kept as simple as possible to speed up the data collection procedure.}

\item{The simulator needs to include an indoor environment (i.e. floor, wall, ceiling, obstacles) with collision and visual properties.}

\item{The simulator needs to be flexible such that parameters can be adjusted in order to investigate their individual impact on performance.}

\end{enumerate}

Table~\ref{tab:solution} summarizes how we addressed these challenges.

\begin{longtable}{c m{11cm}}
\caption{Solutions to challenges regarding the simulation program}
\label{tab:solution} \\
\hline
\hline
 & \multicolumn{1}{c}{\textbf{Solutions}} \\
\hline
\textbf{Challenge 1} & 
\begin{itemize}
\item{We created a simple but accurate URDF model of the DJI Tello drone (Section~\ref{sec:drone})}
\item{We implemented the image capturing behaviour of the DJI Tello drone (Section~\ref{sec:camera})}
\item{We implemented a selection of sensor readings of the DJI Tello drone (Section~\ref{sec:sensors})}
\item{We implemented the control commands of the DJI Tello Drone (Section~\ref{sec:movement})}
\end{itemize}
\\
\textbf{Challenge 2} &
\begin{itemize}
\item{We developed a data collection procedure that produces a labelled dataset which is balanced according to how often the flight commands are typically executed during flight (Section~\ref{sec:data_col})}
\item{We implemented a path planning algorithm that determines the labels of the data samples by calculating the optimal flight path towards the landing platform (Section~\ref{sec:bfs})}
\end{itemize}
\\
\textbf{Challenge 3} &
\begin{itemize}
\item{We used a simple implementation of the control commands (Section~\ref{sec:simple}) to speed up the collection of data samples}
\item{We optimised the path planning algorithm to reduce its Big O complexity (Section~\ref{sec:bfs})}
\end{itemize}
\\
\textbf{Challenge 4} &
\begin{itemize}
\item{We created a simulation environment representing a room which includes a landing platform and cuboid shaped obstacles (Section~\ref{sec:environment})}
\end{itemize}
\\
\textbf{Challenge 5} &
\begin{itemize}
\item{We implemented objects in simulation such that the position and orientation can be individually adjusted (Section~\ref{sec:architecture})}
\item{We conducted experiments in various simulation environments to analyse the impact of environmental changes (Section~\ref{sec:test_flights})}
\end{itemize}
\\
\hline
\hline
\end{longtable}

\citet{polvara} also used a simulation program for training their model. It is not mentioned in their paper whether they developed the simulation program themselves or whether they used an existing simulator. In contrast to us, their simulation program includes domain randomisation features which allowed them to transfer the learnings into the real world \citep{polvara}. Our simulation program does, however, feature the possibility of generating a labelled dataset which \citet{polvara} did not require as they were using a Reinforcement Learning approach. 

\citet{kim} and \citet{ram} did not use a simulation program, but rather manually collected labelled images in the real world.

\section{Architecture}
\label{sec:architecture}

The drone simulation program which has been developed for this project is based on PyBullet \citep{pybullet}. PyBullet is a Python module which wraps the underlying Bullet physics engine through the Bullet C-API \citep{pybullet}. PyBullet \citep{pybullet} has been chosen for this project because it integrates well with TensorFlow \citep{tensorflow}, is well documented, has an active community, is open-source and its physics engine has a proven track record. 

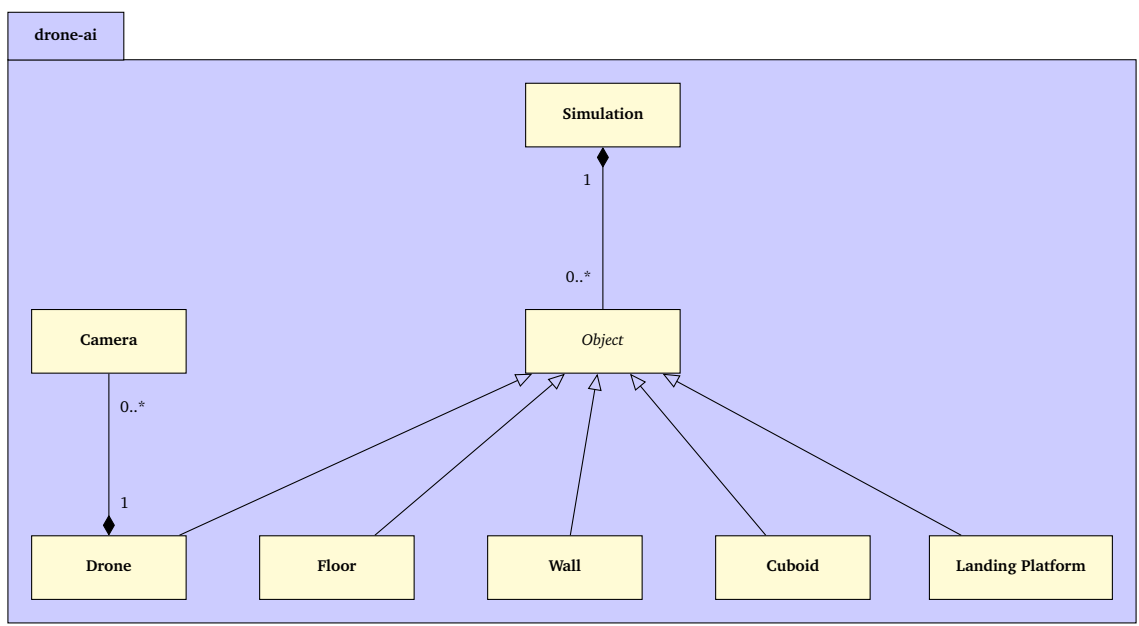
\begin{figure}[!htb]
\centering
\begin{tikzpicture}
\begin{umlpackage}{drone-ai}

\umlsimpleclass[x=6.5, y=9]
{Simulation}

\umlsimpleclass[x=6.5, y=6, type=abstract]
{Object}

\umlsimpleclass[x=0, y=3]
{Drone}

\umlsimpleclass[x=3, y=3]
{Floor}

\umlsimpleclass[x=6, y=3]
{Wall}

\umlsimpleclass[x=9, y=3]
{Cuboid}

\umlsimpleclass[x=12, y=3]
{Landing Platform}

\umlsimpleclass[x=0, y=6]
{Camera}

\umlcompo[mult1=1, mult2=0..*]{Simulation}{Object}

\umlcompo[mult1=1, mult2=0..*]{Drone}{Camera}

\umlinherit{Drone}{Object}
\umlinherit{Floor}{Object}
\umlinherit{Wall}{Object}
\umlinherit{Cuboid}{Object}
\umlinherit{Landing Platform}{Object}

\end{umlpackage}
\end{tikzpicture}
\caption{Class diagram of simulation program}
\label{fig:class_diagram}
\end{figure}

We have integrated the functions provided by the PyBullet packages into an object oriented program written in Python. The program is made up of several classes which are depicted in the class diagram in Figure~\ref{fig:class_diagram}.

The \texttt{Simulation} class manages the execution of the simulation. PyBullet provides both, a \texttt{GUI} mode which creates a graphical user interface (GUI) with 3D OpenGL and allows users to interact with the program through buttons and sliders, as well as a \texttt{DIRECT} mode which sends commands directly to the physics engine without rendering a GUI \citep{pybullet}. \texttt{DIRECT} mode is quicker by two orders of magnitude and therefore more suitable for training a machine learning model. \texttt{GUI} mode  is more useful for debugging the program or to visually retrace the flight path of the drone. Users can switch between these two modes by setting the \texttt{render} variable of the \texttt{Simulation} class. In order to create a simulation environment, a user can do so by calling the \texttt{add\textunderscore room} function which will add walls and ceilings to the environment. Furthermore, using the \texttt{add\textunderscore drone}, \texttt{add\textunderscore landing\textunderscore plattform} and \texttt{add\textunderscore cuboids} functions, a user can place a drone, a landing platform and obstacles at specified positions in space. The \texttt{Simulation} class also implements a function for generating a dataset, a function for calculating the shortest path and a function for executing flights that are controlled by a machine learning model. These functions will be discussed in greater detail in Chapter~\ref{chr:ml}. 

The \texttt{Object} class is an abstract class which provides all functions that are commonly shared across all objects in the simulation environment. These functions can be used for retrieving and changing the position and orientation of the object as well as deleting it from the simulation environment. 

The  \texttt{Drone}, \texttt{Floor}, \texttt{Wall}, \texttt{Cuboid} and \texttt{Landing Platform} classes all inherit from the  \texttt{Object} class and differentiate in regards to their collision and visual properties. 

The \texttt{Drone} class furthermore implements a subset of the original DJI Tello drone commands. These include commands for navigating the drone and commands for reading sensor information. When executing a navigation command, the program simulates the movement of the drone which will be covered in greater detail in Section~\ref{sec:movement}. 

The \texttt{Camera} class can be used to capture images. One or multiple cameras can be attached to a drone and it is possible to specify its resolution, camera angle and field of view properties. The workings of capturing an image in simulation will be explained in Section~\ref{sec:camera}.

\section{Drone}
\label{sec:drone}

The DJI Tello drone has been chosen for this project because it is purpose-built for educational use and is shipped with a software development kit (SDK) which enables a user to connect to the drone via a wifi UDP port, send control commands to the drone and receive state and sensor information as well as a video stream from the drone \citep{sdk}. The flight control system of the DJI Tello drone can translate control commands into rotor movements and thus enables the drone to execute the desired movement. To autonomously navigate the drone, an autonomous system can send control commands to the flight controller of the drone, and thus avoids the need to directly interact with the rotors of the drone. This simplifies the task as the machine learning model doesn't need to learn \textit{how to fly}. The construction of the autonomous system and details of the machine learning model will be covered in Chapter \ref{chr:ml}. 

\begin{figure}[!htb]
    \centering
    \begin{minipage}{.5\textwidth}
        \centering
        \includegraphics[width=0.5\linewidth]{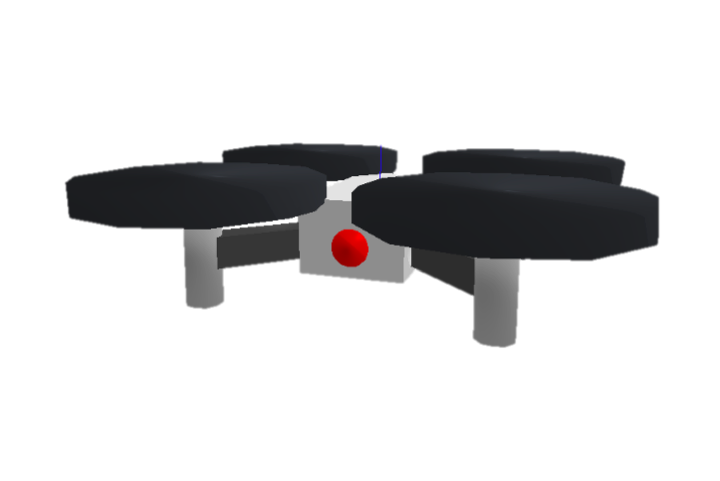}
        \caption{Rendered URDF model}
        \label{fig:tello_urdf}
    \end{minipage}%
    \begin{minipage}{0.5\textwidth}
        \centering
        \includegraphics[width=0.5\linewidth]{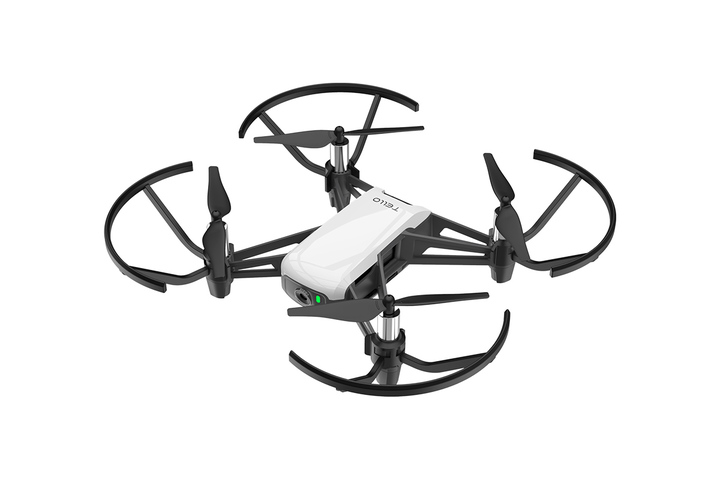}
        \caption{Physical DJI Tello drone (taken from \citep{djiimg})}
        \label{fig:tello_real}
    \end{minipage}
\end{figure}

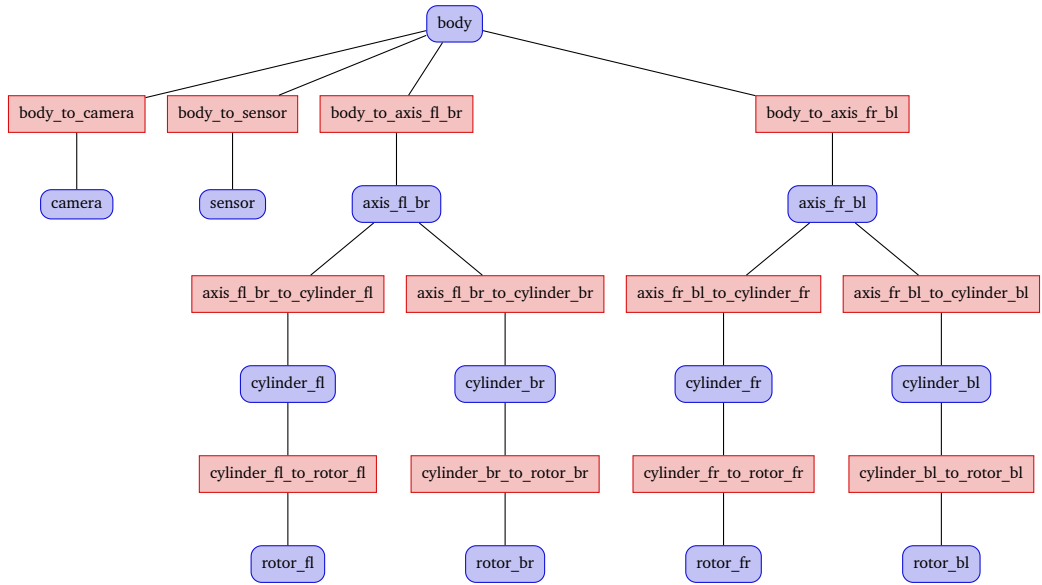
\begin{figure}[!htb]
\centering
\begin{forest}
for tree={
	font=\tiny, 
	joint/.style={rectangle, fill=red!80!darkgray!25, draw=red!80!darkgray},
	link/.style={rectangle, fill=blue!80!darkgray!25, draw=blue!80!darkgray, rounded corners},
}
[body, link
	[body\_to\_camera, joint
		[camera, link]
	]
	[body\_to\_sensor, joint
		[sensor, link]
	]
	[body\_to\_axis\_fl\_br, joint
		[axis\_fl\_br, link
			[axis\_fl\_br\_to\_cylinder\_fl, joint
				[cylinder\_fl, link
					[cylinder\_fl\_to\_rotor\_fl, joint
						[rotor\_fl, link]
					]
				]
			]
			[axis\_fl\_br\_to\_cylinder\_br, joint
				[cylinder\_br, link
					[cylinder\_br\_to\_rotor\_br, joint
						[rotor\_br, link]
					]
				]
			]
		]
	]
	[body\_to\_axis\_fr\_bl, joint
		[axis\_fr\_bl, link
			[axis\_fr\_bl\_to\_cylinder\_fr, joint
				[cylinder\_fr, link
					[cylinder\_fr\_to\_rotor\_fr, joint
						[rotor\_fr, link]
					]
				]
			]
			[axis\_fr\_bl\_to\_cylinder\_bl, joint
				[cylinder\_bl, link
					[cylinder\_bl\_to\_rotor\_bl, joint
						[rotor\_bl, link]
					]
				]
			]
		]
	]
]
\end{forest}
\caption{Hierarchical diagram of links (red) and joints (blue) of DJI Tello URDF model}
\label{fig:urdf_diagram}
\end{figure}

The simulation program contains a model of the DJI Tello drone which we have build in the Unified Robot Description Format (URDF). Figure~\ref{fig:tello_urdf} shows a rendering of the URDF model and can be compared to an image of the physical drone in Figure~\ref{fig:tello_real}. The URDF model has been designed to replicate the dimensions of the drone as closely as possible whilst keeping it simple enough for computational efficiency. The hierarchical diagram depicted in Figure~\ref{fig:urdf_diagram} illustrates the interconnections of the URDF model's links and joints. It includes a specific link to which the camera can be attached. This link is placed at the front of the drone, just like the camera on the real DJI Tello, and is highlighted in red in Figure~\ref{fig:tello_urdf}. In the following, we will discuss how images and sensor information are captured in simulation.

\subsection{Camera}
\label{sec:camera}

PyBullet provides a function called \texttt{getCameraImage} which can be used to capture images in simulation \citep{pybullet}. This function takes the extrinsic and intrinsic matrix of the camera as inputs and returns a RGB image, a depth buffer and a segmentation mask buffer \citep{pybullet}. The extrinsic and intrinsic matrices are used to calibrate the camera according to Tsai's Algorithm \citep{tsai}. Let's take a step back in order to better understand the intuition behind the extrinsic and intrinsic camera matrices.

When we capture an image with the drone's camera, the image should resemble what the drone can \textit{see}. We therefore need to consider where the front side of the drone is facing and where the drone is positioned in the simulation environment. Since the drone can rotate and move to other positions in space, the drone has its own coordinate system which is different from the world coordinate system. PyBullet (or more specifically OpenGL) always performs its calculations based on the world coordinate system and we thus need to translate the drone's coordinate system into the world coordinate system. This is the idea behind an extrinsic camera matrix. \citet[p.~329]{tsai} describes it as ``the transformation from 3D object world coordinate system to the camera 3D coordinate system centered at the optical center''. In order to compute the extrinsic camera matrix, which is also called view matrix, we used the function depicted in Listing~\ref{lst:extrinsic_matrix} in Appendix~\ref{app:extr}.

The intrinsic camera matrix, on the other hand, is concerned with the properties of the camera. It is used for ``the transformation from 3D object coordinate in the camera coordinate system to the computer image coordinate" \cite[p.~329]{tsai}. In other words, it projects a three dimensional visual field onto a two dimensional image which is why the intrinsic camera matrix is also referred to as projection matrix. PyBullet provides the function \texttt{computeProjectionMatrixFOV} for computing the intrinsic camera matrix \citep{pybullet}. This function takes several parameters as inputs: Firstly, it takes in the aspect ratio of the image which we have set to 4:3 as this corresponds to the image dimensions of DJI Tello's camera. Secondly, it takes in a near and a far plane distance value. Anything that is closer than the near plane distance or farther away than the far plane distance, is not displayed in the image. We have set these values to 0.05 meters and 5 meters, respectively. Thirdly, it takes in the vertical field of view (VFOV) of the camera. DJI, however, only advertises the diagonal field of view (DFOV) of its cameras and we thus needed to translate it into VFOV. Using the fundamentals of trigonometry we constructed the following equation to translate DFOV into VFOV:

\begin{equation}
\text{VFOV} = 2 * atan(\frac{h}{d} * tan(\text{DFOV} * 0.5))
\end{equation}

where:
\begin{conditions}
\text{VFOV} & vertical field of view in radian\\
\text{DFOB} & diagonal field of view in radian\\
h & image height \\
d & image diagonal \\ 
\end{conditions}

It is sufficient to calculate the intrinsic camera matrix only once at the start of the simulation, since the properties of the camera don't change. The extrinsic matrix, on the other hand, needs to be calculated every time we capture an image, because the drone might have changed its position and orientation.

Images can be captured at any time during the simulation by simply calling the \texttt{get\textunderscore image} function of the \texttt{Drone} class. This function calls the respective matrix calculation functions of its \texttt{Camera} object and returns the image as a list of pixel intensities.

\subsection{Sensors}
\label{sec:sensors}

The DJI Tello drone is a non-GPS drone and features infra-red sensors and an altimeter in addition to its camera \citep{sdk}. Its SDK provides several read commands which can retrieve information captured by its sensors \citep{sdk}. These include the drone's current speed, battery percentage, flight time, height relative to the take off position, surrounding air temperature, attitude, angular acceleration and distance from take off \citep{sdk}. 

Out of these sensor readings, only the height relative to the take off position, the distance from take off and the flight time are relevant for our simulation program. We have implemented these commands in the \texttt{Drone} class. The \texttt{get\textunderscore height} functions returns the vertical distance between the drone's current position and the take off position in centimetres. The \texttt{get\textunderscore tof} function returns the euclidean distance between the drone's current position and the take off position in centimetres. Instead of measuring the flight time, we are counting the number previously executed flight commands when executing a flight in simulation. 

\section{Movement}
\label{sec:movement}

The physical DJI Tello drone can be controlled by sending SDK commands via a UDP connection to the flight controller of the drone \citep{sdk}. For controlling the movement of the drone, DJI provides two different types of commands: control commands and set commands \citep{sdk}. Control commands direct the drone to fly to a given position relative to its current position. For example, executing the command \texttt{forward(100)}, results in the drone flying one meter in the x-direction of the drone's coordinate system. In contrast, set commands attempt to set new sub-parameter values such as the drone's current speed \citep{sdk}. Using set commands, a user can remotely control the drone via four channels. The difference between these two control concepts is the fact that control commands result in a non-interruptible movement towards a specified target position, whereas the set commands enable the remote navigation of the drone, resulting in a movement that can constantly be interrupted. We decided on using control commands for the simulation program since these can be replicated more accurately. 

The control commands have been implemented in two different variations. The first variation is implemented such that only the position of the drone is updated without simulating the velocity of the drone. This makes command executions quicker and is thus more optimal for collecting data samples in simulation. The second variation simulates the linear and angular velocity of the drone. This variation can be used for visualising the movement and replicates how the real DJI Tello drone would fly towards the target position. In the following, both variations will be discussed in more detail. 

\subsection{Simple Implementation}
\label{sec:simple}

Since control commands are non-interruptible and direct the drone to fly from its current position to a relative target position, the most simple implementation is to only update the drone's position without simulating its movement. In other words, if the drone is advised to fly 20cm forward, we calculate the position to which the drone should fly and set the position of the drone accordingly. This causes the drone to disappear from its current position and reappear at the target position. The DJI Tello SDK provides both, linear movement commands (takeoff, land, up, down, forward, backward, left, right) as well as angular movement commands (cw, ccw) \citep{sdk}. Executing the cw command causes the drone to rotate clock-wise and executing the ccw command causes the drone to rotate counter-clockwise by a given degree. In order to simulate these two commands, it is necessary to update the drone's orientation instead of its position. 

The above mentioned logic has been implemented in the simulation program in the following way: Each control command has its own function in the \texttt{Drone} class. When calling a linear control command, the user needs to provide an integer value corresponding to the anticipated distance in cm that the drone should fly in the desired direction. In a next step, it necessary to check whether the flight path towards the target position is blocked by an obstacle. This is done by the \texttt{path\textunderscore clear} function of the \texttt{Drone} class which uses PyBullet's \texttt{rayTestBatch} function in order to test the flight path for collision properties of other objects in the simulation environment. If the flight path is found to be obstructed, the control command function returns \texttt{False}, implying that the drone would have crashed when executing the control command. If the flight path is clear, the function \texttt{set\textunderscore position\textunderscore relative} of the \texttt{Drone} class is called. This function translates the target position from the drone's coordinate system into the world coordinate system and updates the position of the drone accordingly. 

The execution of angular control commands works in a similar fashion. The functions take in an integer corresponding to the degree by which the drone should rotate in the desired direction. There is no need for collision checking, since the drone doesn't change its position when executing a rotation command. We can thus directly call the \texttt{set\textunderscore orientation\textunderscore relative} function which translates the target orientation into the world coordinate system and updates the simulation.

The simple control command implementation is used for this project since it is less computationally expensive than simulating the actual movement and thus more suitable for the analysis conducted in the following chapters. We nevertheless also implemented a second variation of the flight commands which simulates the linear and angular velocity of the drone. This implementation can be used for future projects that want to use the simulator for other purposes.

\newpage

\subsection{Velocity Implementation}

When creating a \texttt{drone} object, a user has the option to specify whether the velocity of the drone should be simulated. In this case, the control commands execute a different set of private functions which are capable of simulating the linear and angular velocity of the drone. For clarification, these functions do not simulate the physical forces that are acting on the drone but rather the flight path and speed of the drone. It is assumed that the linear and angular velocity of the drone at receipt of a control command are both equal to zero, since control commands on the real DJI Tello drone can only be executed when the drone is hovering in the air. Let's again discuss the implementation of the linear and angular control commands separately beginning with the linear control commands.

If one of the linear control commands is called, the private function \texttt{fly} is executed which takes the target position in terms of the drone's coordinate system as input. This function transforms the target position into the world coordinate system and calculates the euclidean distance from the target position as well as the distance required to accelerate the drone to it's maximum velocity. With these parameters, a callback (called partial in Python) to the \texttt{simulate\textunderscore linear\textunderscore velocity} function is created and set as a variable of the \texttt{drone} object. Once this has been done, everything is prepared to simulate the linear velocity of the drone that results from the linear control command. If the simulation proceeds to the next time step, the callback function is executed which updates the linear velocity of the drone. When calculating the velocity of a movement, it is assumed that the drone accelerates with constant acceleration until the maximum velocity of the drone is reached. Deceleration at the end of the movement is similar. If the flight path is to short in order to reach the maximum velocity, the move profile follows a triangular shape as depicted in Figure~\ref{fig:triangular_profile}. If the flight path is long enough in order to reach the maximum velocity, the velocity stays constant until the deceleration begins, resulting in a trapezoidal move profile (see Figure~\ref{fig:trapezoidal_profile}). 

\begin{figure}[!htb]
    \centering
    \begin{minipage}{.5\textwidth}
	\begin{tikzpicture}[scale=0.8]
\begin{axis}[xtick=\empty, ytick={0, 1}, yticklabels={0, $v_{max}$}, xlabel={Time}, ylabel={Velocity}, xmin=0, xmax=3, ymin=0, ymax=1, enlargelimits=true]
\addplot+[sharp plot] coordinates
{(0,0) (0.75,0.75) (1.5, 0)};
\end{axis}
\end{tikzpicture}
\caption{Triangular move profile}
\label{fig:triangular_profile}
    \end{minipage}%
    \begin{minipage}{0.5\textwidth}
    \centering
\begin{tikzpicture}[scale=0.8]
\begin{axis}[xtick=\empty, ytick={0, 1}, yticklabels={0, $v_{max}$}, xlabel={Time}, ylabel={Velocity}]
\addplot+[sharp plot] coordinates
{(0,0) (1,1) (2,1) (3, 0)};
\end{axis}
\end{tikzpicture}
\caption{Trapezoidal move profile}
\label{fig:trapezoidal_profile}
    \end{minipage}
\end{figure}

A very similar approach has been chosen for the simulation of angular control commands. When the cw or ccw commands are called, the \texttt{rotation} function is executed which takes the target orientation in terms of the drone's coordinate system as input. This function transforms the target orientation into the world coordinate system and calculates the total displacement in radian as well as the displacement required to accelerate the drone. With this information, a callback to the \texttt{simulate\textunderscore angular\textunderscore velocity} function is created and set as an object variable. When stepping through the simulation, the callback function is called which updates the angular velocity of the drone. The move profile of the angular velocity also follows a triangular or trapezoidal shape.

\section{Environment}
\label{sec:environment}

\begin{figure}[!htb]
\centering
\includegraphics[width=0.15\linewidth]{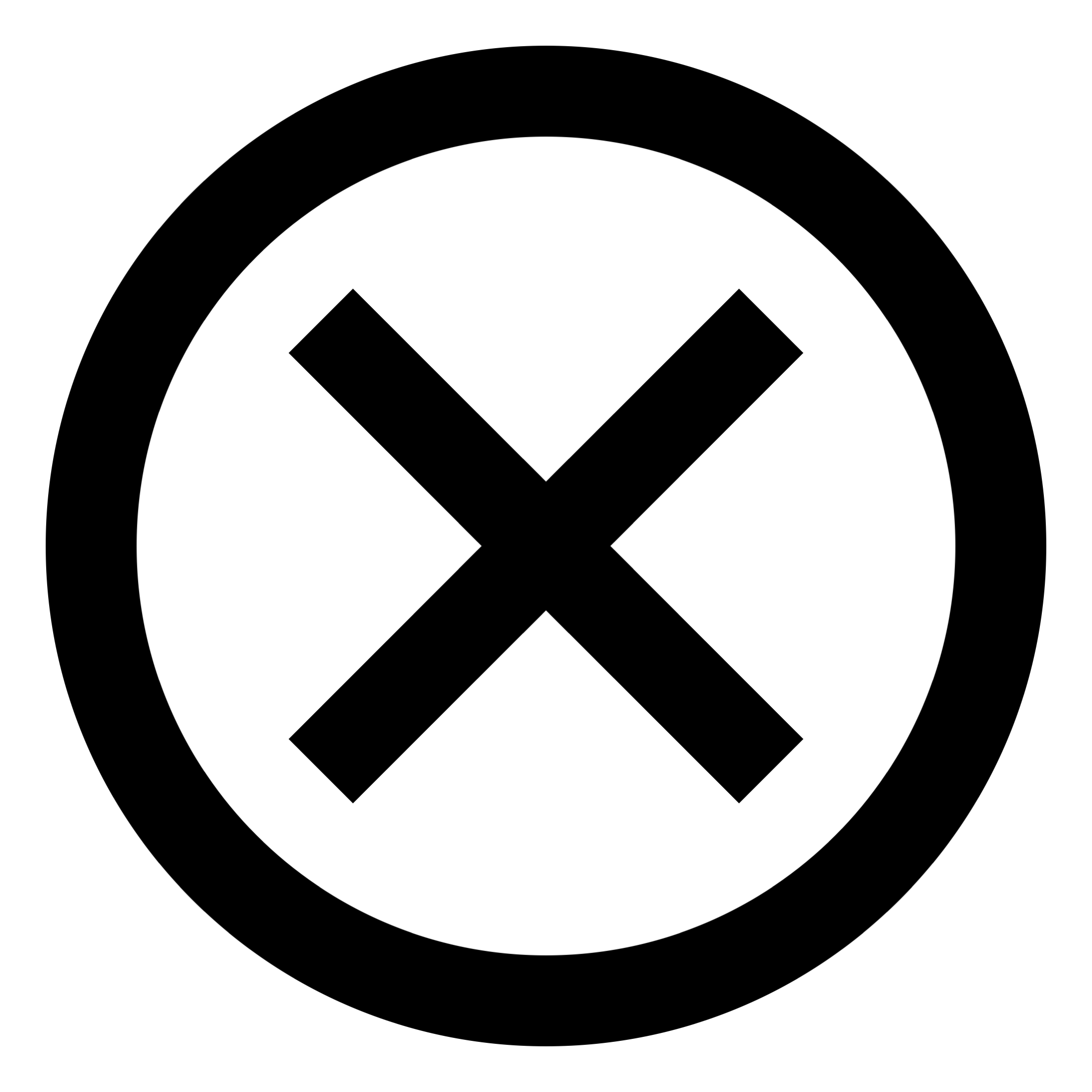}
\caption{Landing platform}
\label{fig:landing_platform}
\end{figure}

The simulation environment is a 3.3m x 3.3m x 2.5m room made up of a floor, four walls and a ceiling. The room can be added to the simulation by calling the \texttt{add\textunderscore room} function in the \texttt{Simulation} class. Inside the room, the drone and the landing platform can be placed. The landing platform measures 60cm x 60cm and is created using a URDF model which references a mesh object. The design of the landing platform is depicted in Figure~\ref{fig:landing_platform} and is similar to the one used by \citet{polvara}. 

\begin{figure}[!htb]
\centering
\includegraphics[width=0.8\linewidth]{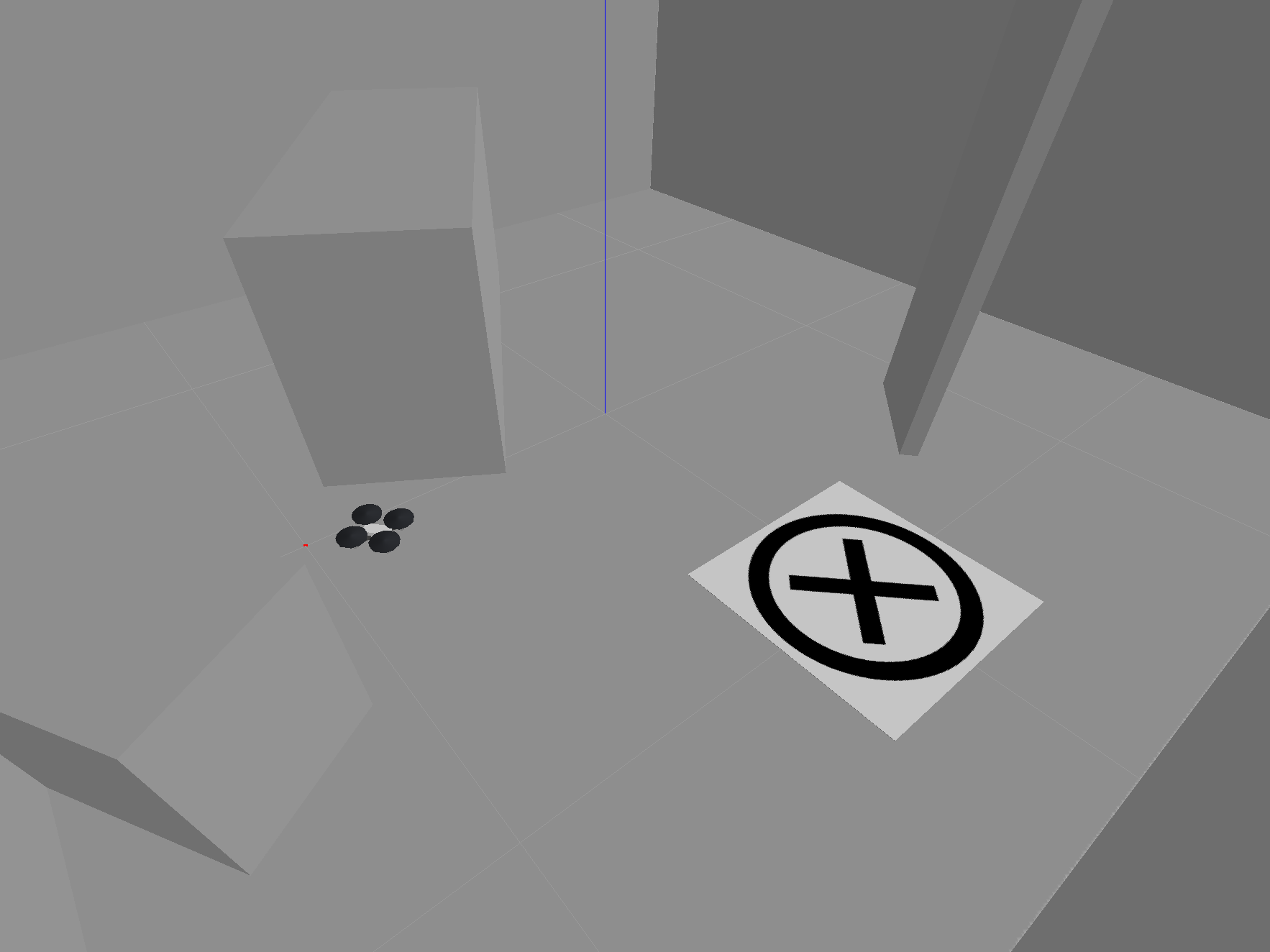}
\caption{Example of a simulation environment}
\label{fig:environment}
\end{figure}

In addition to the drone and the landing platform, obstacles can be placed in the simulation environment. The obstacles are modelled as cuboids and one can specify their position, orientation, width, length, height and color. Obstructed simulation environments can be generated manually, by adding the obstacles one by one, or automatically, in which case the dimensions, positions and orientations are selected using a pseudo-random uniform distribution. A sample simulation environment is depicted in Figure~\ref{fig:environment}.

\newpage

\section{Summary}

This chapter gave an overview of the simulation program that has specifically been developed for this project. It can be used for simulating the control commands and subsequent flight behaviour of the DJI Tello drone. The drone is represented by a URDF model in simulation which replicates the dimensions of the real DJI Tello drone. Furthermore, the drone's camera set-up as well as a selection of sensors has been added for capturing images and reading sensor information during flight. In order to simulate the execution of flight commands, we have introduced a simple but computationally less expensive implementation as well as an implementation in which linear and angular velocities are simulated. The simulation environment consists of a room to which a drone, a landing platform and obstacles can be added. The program is used for the analysis conducted in the following chapters.

\chapter{Machine Learning Model}
\label{chr:ml}

This chapter gives an overview of the machine learning model that has been developed for this project. The task which this model attempts to solve is to autonomously navigate a drone towards a visually marked landing platform.

We used a Supervised Learning approach in order to solve this task. Supervised Learning requires human supervision which comes in the form of a labelled dataset that is used to train the machine learning model \citep{handson}. The labels of the dataset constitute the desired outcome that should be predicted by the model. There are several machine learning systems which can be applied for Supervised Learning such as Decision Trees, Support Vector Machines and Neural Networks. For this project a CNN is used as this is the state of the art machine learning system for image classification \citep{handson}. 

This chapter is divided into three main parts. The first part describes the data collection procedure which is used to generate labelled datasets. The second part identifies the optimal camera set-up for solving the task. And the third part discusses the architectural design of the CNN as well as the optimisation techniques that have been applied in order to find this design. The parts are logically structured and build up on top of each other. Each part solves individual challenges, but it is the combination of all three parts which can be regarded as a coherent approach of solving the task at hand.

\section{Dataset}
\label{sec:dataset}

A dataset that is suitable for this project needs to fulfil very specific requirements. Each data sample needs to include multiple features and one label. The features correspond the state of the drone which is specified by the visual input of the drone's camera and the sensor information captured by its sensors. The label corresponds to the desired flight command which the drone should execute from this state in order to fly towards the landing platform. As a labelled dataset which fulfils these requirements doesn't already exist, a data collection procedure had to be developed that is capable of generating suitable data samples. To collect these data samples we use our simulation environment which has been introduced in Chapter~\ref{chr:simulation}. 

In the following, we will in a first step describe the workings of the data collection procedure and in a second step elaborate on the algorithm that is applied for finding an optimal flight path. 

\subsection{Data Collection Procedure}
\label{sec:data_col}

\begin{algorithm}[!htb]
\caption{Naive data collection}
\label{alg:naive_data_col}
\begin{algorithmic}[1]
\Procedure{generate dataset}{size}
\State dataset = [\:]
\State landing = LandingPlatform()
\State drone = Drone()
\State count = 0
\While{count $<$ size}
	\State landing.set\_position\_and\_orientation\_3d\_random()
	\State drone.set\_position\_and\_orientation\_2d\_random()
	\State image = drone.get\_image()
	\State sensor = drone.get\_sensor()
	\State label = calculate\_optimal\_command()
	\State sample = [image, sensor, label]
	\State dataset.append(sample)
	\State count++
\EndWhile
\Return dataset
\EndProcedure
\end{algorithmic}
\end{algorithm}

When creating a new algorithm a good practice is to begin with a very simple algorithm and expand its complexity depending on the challenges you encounter. A naive data collection procedure is depicted in Algorithm~\ref{alg:naive_data_col} which works as follows: At first, the algorithm places the drone and the landing platform at random positions in the simulation environment. The position of the drone can be any position in the three dimensional simulation space and the landing platform is placed somewhere on the ground. The algorithm then captures an image using the drone's camera and requests sensor information from the drone's sensors. In a next step, the desired flight command, which should be executed by the drone in order to fly towards the landing platform, is computed. The desired flight command can be derived from the optimal flight path which can be calculated using a breadth first search (BFS) algorithm since both, the position of the drone and the position of the landing platform, are known in simulation. The workings of the BFS algorithm will be discussed later in Section~\ref{sec:bfs}. Finally, the image and sensor information, which make up the features of the data sample, are stored together with the desired flight command, which corresponds to the label of the data sample. Repeating this procedure generates a dataset. 

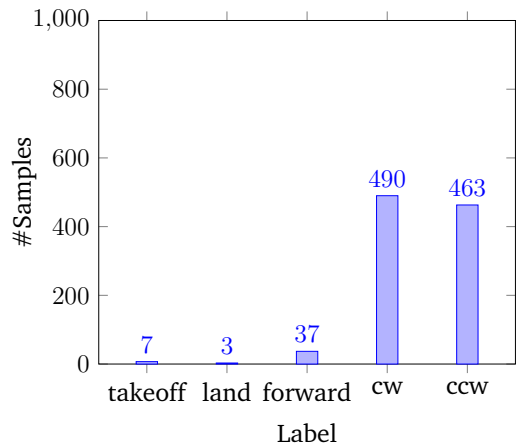
\begin{figure}[!htb]
\centering
\begin{tikzpicture}[scale=0.8]
\begin{axis}
[
xlabel={Label}, 
xlabel style={yshift=-0.3cm},
xtick=data,
xticklabel style={rotate=0},
symbolic x coords={takeoff, land, forward, cw, ccw},
ylabel={\#Samples},
ylabel style={yshift=0cm},
ymin=0, 
ymax=1000, 
enlarge x limits=0.15,
nodes near coords, 
nodes near coords align={vertical}, 
ybar,  
]
\addplot coordinates {(takeoff,7) (land,3) (forward,37) (cw,490) (ccw,463) };
\end{axis}
\end{tikzpicture}
\caption{Histogram of 1\,000 data sample labels produced by the naive data collection procedure in the non-obstructed simulation environment}
\label{fig:hist_naive}
\end{figure}

A downside of this procedure is, however, that it produces a very unbalanced dataset. As depicted in Figure~\ref{fig:hist_naive}, the desired flight command is almost always to rotate clockwise (cw) or counterclockwise (ccw). This makes sense, because we allow the drone to only execute one of the following five commands: takeoff, land, fly forward 20cm, rotate cw 10\textdegree{} and rotate ccw 10\textdegree. Horizontal movement is therefore limited to only flying forward because this is where the camera of the drone is facing. In order to fly left, the drone first needs to rotate ccw multiple times and then fly forward. Since the naive data collection procedure places the drone at a random position and with a random orientation, it is most likely not facing towards the landing platform. The desired flight command is thus in most cases to rotate towards the landing platform which results in an unbalanced dataset. 

In order to overcome this unbalanced data problem, a more sophisticated data collection procedure has been developed for which the pseudo-code is shown in Algorithm~\ref{alg:soph_data_col}. This algorithm begins with placing both, the drone and the landing platform, at random positions on the ground of the simulation environment using a uniform distribution. Notice that the drone is now placed on the ground which is different to the naive algorithm that places it somewhere in space. It then calculates the optimal flight path using a BFS algorithm. This path includes a series of commands which the drone needs to execute in order to fly from its current position to the landing platform. Beginning with the drone's starting position, the algorithm generates a data sample by capturing and storing the image and sensor information as well as the command that should be executed next. It then advises the drone to execute this command in simulation which consequently repositions the drone. Using the new state of the drone we can again capture camera and sensor information and hereby generate a new data sample. These steps are repeated until the drone has reached the landing platform. Once the drone has landed, both the drone and the landing platform are repositioned and a new shortest path is calculated. Repeating this procedure multiple times generates a dataset.

\begin{algorithm}[!htb]
\caption{Sophisticated data collection}
\label{alg:soph_data_col}
\begin{algorithmic}[1]
\Procedure{generate dataset}{size}
\State dataset = [\:]
\State landing = LandingPlatform()
\State drone = Drone()
\State count = 0
\While{count $<$ size}
	\State landing.set\_position\_and\_orientation\_2d\_random()
	\State drone.set\_position\_and\_orientation\_2d\_random()
	\State path = caclulate\_shortest\_path()
	\ForEach{command \textbf{in} path}
		\If{count $>=$ size}
			\Break
		\EndIf
		\State image = drone.get\_image()
		\State sensor = drone.get\_sensor()
		\State label = command
		\State sample = [image, sensor, label]
		\State dataset.append(sample)
		\State drone.execute(command)
		\State count++
	\EndFor
\EndWhile
\Return dataset
\EndProcedure
\end{algorithmic}
\end{algorithm}

\begin{figure}[!htb]
\centering
\begin{tikzpicture}[scale=0.8]
\begin{axis}
[
xlabel={Label}, 
xlabel style={yshift=-0.3cm},
xtick=data,
xticklabel style={rotate=0},
symbolic x coords={takeoff, land, forward, cw, ccw},
ylabel={\#Samples},
ylabel style={yshift=0cm},
ymin=0, 
ymax=1000, 
enlarge x limits=0.15,
nodes near coords, 
nodes near coords align={vertical}, 
ybar,  
]
\addplot coordinates {(takeoff,54) (land,53) (forward,396) (cw,246) (ccw,251) };
\end{axis}
\end{tikzpicture}
\caption{Histogram of 1\,000 data sample labels produced by the sophisticated data collection procedure in the non-obstructed simulation environment}
\label{fig:hist_soph}
\end{figure}
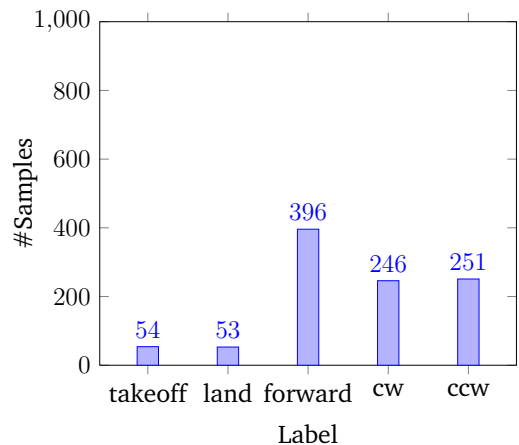

As shown in Figure~\ref{fig:hist_soph}, the dataset is now significantly more balanced compared to the one produced by the naive data collection procedure. It includes sample labels of all five flight commands and its label distribution reflects how often the commands are executed during a typical flight in the simulated environment. Since the dataset, on which Figure~\ref{fig:hist_soph} is based on, includes 54 flights we can deduct that the average flight path includes one takeoff command, nine rotational commands (cw or ccw), seven forward commands and one land command. Using the sophisticated data collection procedure it is thus possible to generate a dataset which is balanced according to the number of times a drone usually executes the commands during flight. As the dataset can be of arbitrary size it is also possible to adapt the label distribution such that every label is represented equally in the dataset. We did, however, not apply this technique and instead decided on using class weights for the loss function of the neural network. More on this in Section~\ref{sec:foundations}.

\subsection{Optimal Flight Path}
\label{sec:bfs}

The above mentioned data collection algorithms require information about the optimal flight path. The optimal flight path is the one that requires the drone to execute the least number of flight commands in order to fly from its current position to the landing platform. Since both, the position of the drone and the position of the landing platform, are known in simulation, it is possible to compute the optimal flight path.

In order to find the optimal flight path it was necessary to develop an algorithm that is specifically targeted at our use case. The general idea behind this algorithm is to construct a graph that consists of nodes, representing the positions that the drone can reach, and links, representing the flight commands that the drone can execute. Using a BFS algorithm we can then compute the shortest path between any two nodes of the graph. Since the shortest path represents the links that need to be traversed in order to get from the start node to the target node, we can thus derive the series of commands that the drone needs to execute in order to fly from the position represented by the start node to the position represented by the target node. This path is also the optimal path as it includes the least number of flight commands.

Let's translate this general idea into our specific use case. We can begin with the construction of the graph by adding the drone's current pose as the central node. The pose of an object is the combination of its position and orientation. It is necessary to use the drone's pose and not simply its position because the drone can rotate along its z-axis. From this starting pose, the drone can reach a new pose by executing one of its five flight commands. We can therefore add five new nodes that represent the five poses that can be reached from the drone's current pose and connect them using links which have the label of the respective flight command. From each of the new poses, the drone can again execute five new flight commands which adds new poses to the graph. Repeating this procedure, a graph is constructed which includes all the poses that the drone can reach from its current pose. Since we are searching for the shortest path between the drone and the landing platform, we can stop adding new poses to the graph once a pose has been found which is within a certain threshold of the position of the landing platform. Note that for the breaking condition only the position of the drone and not its orientation is being considered, as it doesn't matter in which direction the drone is facing when it lands on the platform. When we add the nodes to the graph in a BFS fashion we can ensure that we have found the shortest path, once a pose is within the landing area.

One problem with this approach is that the graph is growing way too quickly resulting in too many poses. The upper bound time complexity of a BFS algorithm is $\mathcal{O}(b^d)$, with branching factor $b$ and search tree depth $d$ \citep{russel}. Since a drone can execute five different commands from every pose, the branching factor, meaning the number of outgoing links of every node, is equal to five. The depth on the other hand depends on the distance between the drone's current pose and the landing platform. If, for example, the landing platform is at minimum 20 flight commands away from the drone, it would require up to $5^{20}$ iterations to find the shortest path using this algorithm. This is too computationally expensive and it was thus necessary to optimise the BFS algorithm. 

\begin{algorithm}[!htb]
\caption{Optimal flight path (optimised BFS)}
\label{alg:bfs}
\begin{algorithmic}[1]
\Procedure{get\_optimal\_flight\_path}{drone, landing}
\State queue = [\:]	\Comment{FIFO queue data structure}
\State visited = \{\:\} \Comment{Dictionary data structure}
\State path = [\:] \Comment{List data structure}
\State start\_pos, start\_ori = drone.get\_position\_and\_orientation()
\State landing\_pos = landing.get\_position()
\State queue.enqueue([start\_pos, start\_ori, path])
\While{queue.length $>$ 0}
	\State pos, ori, path = queue.dequeue()
	\State drone.set\_position\_and\_orientation(pos, ori)
	\If{euclidean\_distance(pos, landing\_pos) $<$ 10cm} \Comment{Break condition}
		\State drone.set\_position\_and\_orientation(start\_pos, start\_ori)
		\Return path
	\EndIf
	\For{command \textbf{in} [takeoff, land, forward, cw, ccw]}
		\If{drone.execute(command)} \Comment{Returns false if drone crashes}
			\State new\_pos, new\_ori = drone.get\_position\_and\_orientation()
			\State square = get\_square\_of\_position(pos)
			\State yaw = get\_rounded\_yaw\_of\_orientation(ori)
			\If{[square, yaw] \textbf{not in} visited}
				\State visted[square, yaw] = true
				\State new\_path = path.append(command) \Comment{Deep copy}
				\State queue.enqueue([new\_pos, new\_ori, new\_path])
			\EndIf
			\State drone.set\_position\_and\_orientation(pos, ori)
		\EndIf
	\EndFor
\EndWhile
\Return path
\EndProcedure
\end{algorithmic}
\end{algorithm}

Algorithm~\ref{alg:bfs} shows the pseudo-code of the optimised BFS algorithm. Three main optimisation features have been implemented in order to speed up the algorithm. 

\begin{itemize}
\item We make use of efficient data structures. A FIFO queue is used to store the nodes that need to be visited next. Both queue operations, \texttt{enqueue} and \texttt{dequeue} have a time complexity of $\mathcal{O}(1)$ \citep{cormen}. A dictionary is used to store the nodes that have already been visited. Since the dictionary is implemented using a hash table, we can check whether a node is already included in the dictionary (Line 20) with time complexity $\mathcal{O}(1)$ \citep{cormen}. A list is used to keep track of the \texttt{path}, to which a new flight command can be appended with time complexity $\mathcal{O}(1)$ \citep{cormen}.
  
\item We check in Line 16 whether the execution of a command is valid. A command is invalid if it would result in a crash (because the trajectory is obstructed by some object) or if the DJI Tello SDK doesn't allow its execution. The takeoff command can for example only be executed when the drone is sitting on a surface. Likewise the forward, cw, ccw and land commands can only be executed when the drone is in the air. If a command is invalid, it is skipped and no pose is added to the queue. This reduces the branching factor because nodes have on average fewer outgoing links. 

\item The most drastic optimisation feature is the partitioning of space. The general idea behind this feature is that it isn't necessary to follow every branch of the BFS tree as most of these branches will lead to similar poses. For example if we find a new pose at depth ten of the BFS tree and we have already found a very similar pose at a lower level of the BFS tree, it is almost certain that the new pose won't help us find the shortest path and can thus be ignored. Our algorithm implements this logic by subdividing the simulation environment into cubes of equal sizes. Each position in the simulation environment is therefore part of one cube. And since we also need the orientation in order to identify the pose of the drone, every cube contains a limited number of yaw angles. Note that only the yaw angle of the drone's orientation is considered as it can only rotate along its z-axis. Every drone pose that is discovered during the search can thus be allocated to one square and one (rounded) yaw angle. The combination of square and yaw angle needs to be stored in the \texttt{visited} dictionary and if an entry of this combination is already included, we know that a similar pose has already been discovered. We also know that the previously discovered pose is fewer links away from the starting node and is therefore a better candidate for the shortest path. We can thus ignore a pose if its square and yaw have already been visited. This significantly reduces the time complexity of the algorithm with an upper bound of $\mathcal{O}(c * y)$ where $c$ is the number of cubes and $y$ is the number of yaw angles. Reducing the length of the cube edges (which increases the number of cubes that fill up the environment) results in a worse upper bound time complexity but also increases the chances of finding the true shortest path. Reducing the step size between the yaw angles (which increases the number of yaw angles) has the same effect. On the other hand, if the cubes are too large or there are too few yaw angles, the shortest path might not be found. Since the forward command makes the drone fly 20cm in horizontal direction and the cw/ccw rotation commands rotate the drone by 10\textdegree{}, the cube edges have been set to $\frac{20}{\sqrt{2}}$cm and the yaw angle step size has been set to 10\textdegree{}.
\end{itemize}

The upper bound time complexity of the optimised BFS algorithm, when applied in the simulation environment described in Section~\ref{sec:environment} is therefore:

\begin{equation}  \label{eq:1}
\mathcal{O}(\frac{w * d * h}{(c)^3} * \frac{360\degree }{y})
\end{equation}

where:
\begin{conditions}
w & width of the room \\
d & depth of the room \\   
h & height of the room \\
c & cube edge length \\
y & yaw angle step size  \\
\end{conditions}

Inserting the appropriate numeric values into Equation~\ref{eq:1} reveals that in the worst case, 135\,252 iterations need to be performed to calculate the optimal flight path in our simulation environment. This is over 25 orders of magnitude lower than if we were to use a non-optimised BFS algorithm.

\newpage

\section{Optimal Camera Set-Up}

The DJI Tello drone is shipped with a forward facing camera which is slightly tilted downward at a 10\textdegree{} angle and with a DFOV of 82.6\textdegree{}. We hypothesized that with this camera set-up, (i) the drone would struggle to vertically land on the landing platform (ii) as the landing platform goes out of sight once the drone gets too close. 

Our analysis showed that the original camera set-up of the DJI Tello drone is suboptimal for our task at hand. We found a fisheye set-up with a DFOV of 150\textdegree{} to be optimal. 

In the following, we will in a first step discuss how we tested our hypothesis and in a second step layout alternative camera set-ups.

\subsection{DJI Tello Camera}

In order to test our hypothesis, we generated a dataset with the original DJI Tello camera set-up, used this dataset to train a CNN and evaluated its performance. 

\begin{figure}[!htb]
\centering
\includegraphics[width=\textwidth]{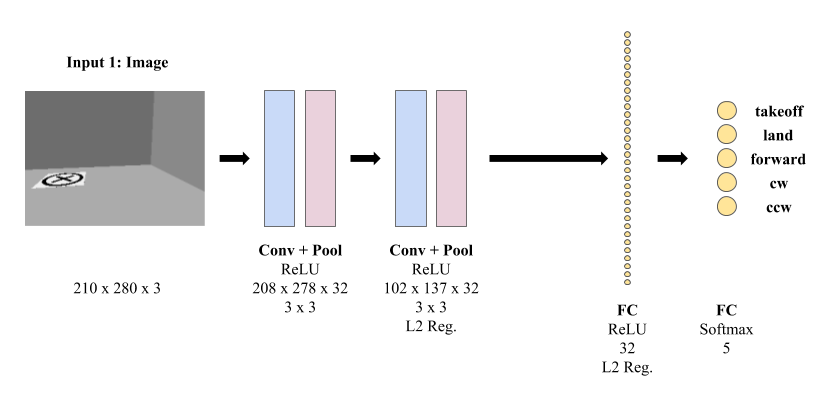}
\caption{CNN architecture for optimal camera set-up analysis}
\label{fig:cnn_simple}
\end{figure}

The dataset includes 10\,000 data samples which have been collected in a non-obstructed simulation environment using the previously described data collection procedure. Each data sample is made up of a 280 x 210 image that has been taken from a forward facing camera (tilted 10\textdegree{} downward with 82.6\textdegree{} DFOV) and a label which corresponds to the desired flight command that should be executed. There is no need for capturing sensor information as we are only focusing on the camera set-up. The dataset has been partitioned into a training and a test set using a 70/30 split. There is no need for a validation set as we don't require hyper parameter tuning in order to test our hypothesis.

As illustrated in Figure~\ref{fig:cnn_simple}, the CNN architecture is made up of an input layer taking in a RGB image, two convolutional layers, two max-pooling layers, one fully connected layer and an output layer with five neurons which correspond to the five commands that the drone can execute. It uses ReLU activation functions, categorical cross entropy loss and Adam optimisation. 

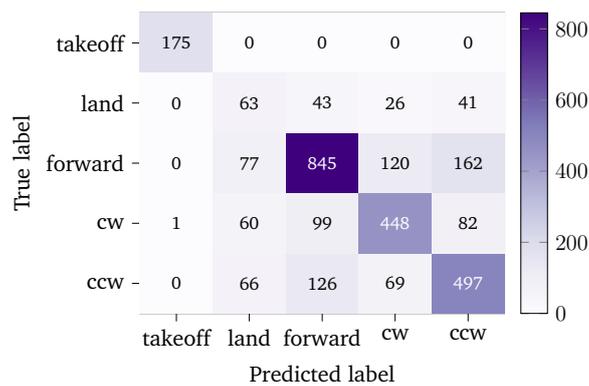
\begin{figure}[!htb]
\centering
\begin{tikzpicture}[scale=0.7]

\begin{axis}[
xlabel style={yshift=-0.3cm},
ylabel style={yshift=1cm},
colorbar,
colorbar style={ylabel={}},
colormap={mymap}{[1pt]
  rgb(0pt)=(0.988235294117647,0.984313725490196,0.992156862745098);
  rgb(1pt)=(0.937254901960784,0.929411764705882,0.96078431372549);
  rgb(2pt)=(0.854901960784314,0.854901960784314,0.92156862745098);
  rgb(3pt)=(0.737254901960784,0.741176470588235,0.862745098039216);
  rgb(4pt)=(0.619607843137255,0.603921568627451,0.784313725490196);
  rgb(5pt)=(0.501960784313725,0.490196078431373,0.729411764705882);
  rgb(6pt)=(0.415686274509804,0.317647058823529,0.63921568627451);
  rgb(7pt)=(0.329411764705882,0.152941176470588,0.56078431372549);
  rgb(8pt)=(0.247058823529412,0,0.490196078431373)
},
point meta max=845,
point meta min=0,
tick align=outside,
tick pos=left,
x grid style={white!69.0196078431373!black},
xlabel={Predicted label},
xmin=-0.5, xmax=4.5,
xtick style={color=black},
xtick={0,1,2,3,4},
xticklabels={takeoff,land,forward,cw,ccw},
y dir=reverse,
y grid style={white!69.0196078431373!black},
ylabel={True label},
ymin=-0.5, ymax=4.5,
ytick style={color=black},
ytick={0,1,2,3,4},
yticklabels={takeoff,land,forward,cw,ccw}
]
\addplot graphics [includegraphics cmd=\pgfimage,xmin=-0.5, xmax=4.5, ymin=4.5, ymax=-0.5] {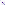};
\draw (axis cs:0,0) node[
  scale=0.789473684210526,
  text=black,
  rotate=0.0
]{175};
\draw (axis cs:1,0) node[
  scale=0.789473684210526,
  text=black,
  rotate=0.0
]{0};
\draw (axis cs:2,0) node[
  scale=0.789473684210526,
  text=black,
  rotate=0.0
]{0};
\draw (axis cs:3,0) node[
  scale=0.789473684210526,
  text=black,
  rotate=0.0
]{0};
\draw (axis cs:4,0) node[
  scale=0.789473684210526,
  text=black,
  rotate=0.0
]{0};
\draw (axis cs:0,1) node[
  scale=0.789473684210526,
  text=black,
  rotate=0.0
]{0};
\draw (axis cs:1,1) node[
  scale=0.789473684210526,
  text=black,
  rotate=0.0
]{63};
\draw (axis cs:2,1) node[
  scale=0.789473684210526,
  text=black,
  rotate=0.0
]{43};
\draw (axis cs:3,1) node[
  scale=0.789473684210526,
  text=black,
  rotate=0.0
]{26};
\draw (axis cs:4,1) node[
  scale=0.789473684210526,
  text=black,
  rotate=0.0
]{41};
\draw (axis cs:0,2) node[
  scale=0.789473684210526,
  text=black,
  rotate=0.0
]{0};
\draw (axis cs:1,2) node[
  scale=0.789473684210526,
  text=black,
  rotate=0.0
]{77};
\draw (axis cs:2,2) node[
  scale=0.789473684210526,
  text=white,
  rotate=0.0
]{845};
\draw (axis cs:3,2) node[
  scale=0.789473684210526,
  text=black,
  rotate=0.0
]{120};
\draw (axis cs:4,2) node[
  scale=0.789473684210526,
  text=black,
  rotate=0.0
]{162};
\draw (axis cs:0,3) node[
  scale=0.789473684210526,
  text=black,
  rotate=0.0
]{1};
\draw (axis cs:1,3) node[
  scale=0.789473684210526,
  text=black,
  rotate=0.0
]{60};
\draw (axis cs:2,3) node[
  scale=0.789473684210526,
  text=black,
  rotate=0.0
]{99};
\draw (axis cs:3,3) node[
  scale=0.789473684210526,
  text=white,
  rotate=0.0
]{448};
\draw (axis cs:4,3) node[
  scale=0.789473684210526,
  text=black,
  rotate=0.0
]{82};
\draw (axis cs:0,4) node[
  scale=0.789473684210526,
  text=black,
  rotate=0.0
]{0};
\draw (axis cs:1,4) node[
  scale=0.789473684210526,
  text=black,
  rotate=0.0
]{66};
\draw (axis cs:2,4) node[
  scale=0.789473684210526,
  text=black,
  rotate=0.0
]{126};
\draw (axis cs:3,4) node[
  scale=0.789473684210526,
  text=black,
  rotate=0.0
]{69};
\draw (axis cs:4,4) node[
  scale=0.789473684210526,
  text=white,
  rotate=0.0
]{497};
\end{axis}

\end{tikzpicture}
\caption{Confusion matrix \textit{front camera} model (67.6\% accuracy) }
\label{fig:conf_matrix_front826}
\end{figure}

The model (hereafter referred to as ``front camera model") has been trained over 20 epochs with a batch size of 32 and achieves a prediction accuracy of 67.6\% on the test set. The confusion matrix plotted in Figure~\ref{fig:conf_matrix_front826} reveals that the performance of the front camera model significantly varies across the desired flight commands. On the one hand, it manages to correctly predict the takeoff, forward and rotation commands in most cases. On the other hand, it misclassifies 63.6\% of data samples that are labelled as ``land". This supports the first part of our hypothesis, because it clearly shows that the front camera model struggles with landing the drone.

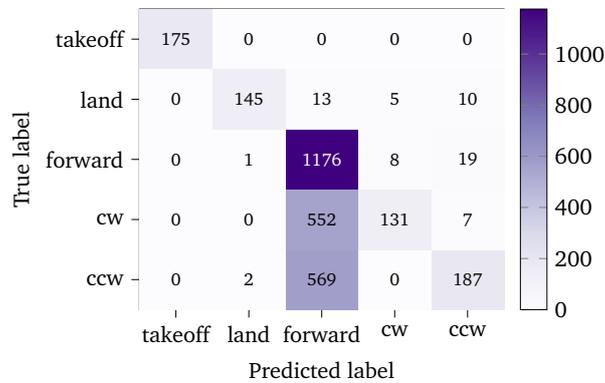
\begin{figure}[!htb]
\centering
\begin{tikzpicture}[scale=0.7]

\begin{axis}[
xlabel style={yshift=-0.3cm},
ylabel style={yshift=1cm},
colorbar,
colorbar style={ytick={0,200,400,600,800,1000},yticklabels={0,200,400,600,800,1000},ylabel={}},
colormap={mymap}{[1pt]
  rgb(0pt)=(0.988235294117647,0.984313725490196,0.992156862745098);
  rgb(1pt)=(0.937254901960784,0.929411764705882,0.96078431372549);
  rgb(2pt)=(0.854901960784314,0.854901960784314,0.92156862745098);
  rgb(3pt)=(0.737254901960784,0.741176470588235,0.862745098039216);
  rgb(4pt)=(0.619607843137255,0.603921568627451,0.784313725490196);
  rgb(5pt)=(0.501960784313725,0.490196078431373,0.729411764705882);
  rgb(6pt)=(0.415686274509804,0.317647058823529,0.63921568627451);
  rgb(7pt)=(0.329411764705882,0.152941176470588,0.56078431372549);
  rgb(8pt)=(0.247058823529412,0,0.490196078431373)
},
point meta max=1176,
point meta min=0,
tick align=outside,
tick pos=left,
x grid style={white!69.0196078431373!black},
xlabel={Predicted label},
xmin=-0.5, xmax=4.5,
xtick style={color=black},
xtick={0,1,2,3,4},
xticklabels={takeoff,land,forward,cw,ccw},
y dir=reverse,
y grid style={white!69.0196078431373!black},
ylabel={True label},
ymin=-0.5, ymax=4.5,
ytick style={color=black},
ytick={0,1,2,3,4},
yticklabels={takeoff,land,forward,cw,ccw}
]
\addplot graphics [includegraphics cmd=\pgfimage,xmin=-0.5, xmax=4.5, ymin=4.5, ymax=-0.5] {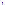};
\draw (axis cs:0,0) node[
  scale=0.789473684210526,
  text=black,
  rotate=0.0
]{175};
\draw (axis cs:1,0) node[
  scale=0.789473684210526,
  text=black,
  rotate=0.0
]{0};
\draw (axis cs:2,0) node[
  scale=0.789473684210526,
  text=black,
  rotate=0.0
]{0};
\draw (axis cs:3,0) node[
  scale=0.789473684210526,
  text=black,
  rotate=0.0
]{0};
\draw (axis cs:4,0) node[
  scale=0.789473684210526,
  text=black,
  rotate=0.0
]{0};
\draw (axis cs:0,1) node[
  scale=0.789473684210526,
  text=black,
  rotate=0.0
]{0};
\draw (axis cs:1,1) node[
  scale=0.789473684210526,
  text=black,
  rotate=0.0
]{145};
\draw (axis cs:2,1) node[
  scale=0.789473684210526,
  text=black,
  rotate=0.0
]{13};
\draw (axis cs:3,1) node[
  scale=0.789473684210526,
  text=black,
  rotate=0.0
]{5};
\draw (axis cs:4,1) node[
  scale=0.789473684210526,
  text=black,
  rotate=0.0
]{10};
\draw (axis cs:0,2) node[
  scale=0.789473684210526,
  text=black,
  rotate=0.0
]{0};
\draw (axis cs:1,2) node[
  scale=0.789473684210526,
  text=black,
  rotate=0.0
]{1};
\draw (axis cs:2,2) node[
  scale=0.789473684210526,
  text=white,
  rotate=0.0
]{1176};
\draw (axis cs:3,2) node[
  scale=0.789473684210526,
  text=black,
  rotate=0.0
]{8};
\draw (axis cs:4,2) node[
  scale=0.789473684210526,
  text=black,
  rotate=0.0
]{19};
\draw (axis cs:0,3) node[
  scale=0.789473684210526,
  text=black,
  rotate=0.0
]{0};
\draw (axis cs:1,3) node[
  scale=0.789473684210526,
  text=black,
  rotate=0.0
]{0};
\draw (axis cs:2,3) node[
  scale=0.789473684210526,
  text=black,
  rotate=0.0
]{552};
\draw (axis cs:3,3) node[
  scale=0.789473684210526,
  text=black,
  rotate=0.0
]{131};
\draw (axis cs:4,3) node[
  scale=0.789473684210526,
  text=black,
  rotate=0.0
]{7};
\draw (axis cs:0,4) node[
  scale=0.789473684210526,
  text=black,
  rotate=0.0
]{0};
\draw (axis cs:1,4) node[
  scale=0.789473684210526,
  text=black,
  rotate=0.0
]{2};
\draw (axis cs:2,4) node[
  scale=0.789473684210526,
  text=black,
  rotate=0.0
]{569};
\draw (axis cs:3,4) node[
  scale=0.789473684210526,
  text=black,
  rotate=0.0
]{0};
\draw (axis cs:4,4) node[
  scale=0.789473684210526,
  text=black,
  rotate=0.0
]{187};
\end{axis}

\end{tikzpicture}
\caption{Confusion matrix \textit{bottom camera} model (60.5\% accuracy) }
\label{fig:conf_matrix_bottom826}
\end{figure}

In order to prove the second part of our hypothesis, we developed a second model (hereafter referred to as ``bottom camera model"). This model uses the same CNN architecture and is trained for the same number of epochs, but the images in the dataset have been taken using a downward facing camera (tilted 90\textdegree{} downward with 82.6\textdegree{} DFOV). In fact, this dataset was created using the same flight paths as the previous dataset which means that the images were taken from the same positions and the label distribution is equivalent in both datasets. The only difference between these two datasets is the camera angle at which the images were taken. The bottom camera model achieves a 60.2\% prediction accuracy on the test set which is worse than the front camera model. But as we can see in the confusion matrix depicted in Figure~\ref{fig:conf_matrix_bottom826}, the bottom camera model correctly predicts 83.8\% of the land commands in the test set. This performance variation makes sense, because with a downward facing camera, the drone can only see the landing platform when it is hovering above it. The bottom camera model therefore knows when to land, but doesn't know how to fly to the landing platform. This supports the second part of our hypothesis, since we have shown that the bad landing performance of the front camera model is due to the fact that the landing platform is out of sight once the drone gets too close to it.

\subsection{Camera Adjustments}

Since neither the front camera model, nor the bottom camera model, are capable of predicting the comprehensive set of flight commands, it was necessary to come up with a new camera set-up. Three different set-ups have been tested and are in the following compared against each other. 

\begin{figure}[!htb]
\centering
\begin{tikzpicture}[scale=0.7]

\begin{axis}[
xlabel style={yshift=-0.3cm},
ylabel style={yshift=1cm},
colorbar,
colorbar style={ylabel={}},
colormap={mymap}{[1pt]
  rgb(0pt)=(0.988235294117647,0.984313725490196,0.992156862745098);
  rgb(1pt)=(0.937254901960784,0.929411764705882,0.96078431372549);
  rgb(2pt)=(0.854901960784314,0.854901960784314,0.92156862745098);
  rgb(3pt)=(0.737254901960784,0.741176470588235,0.862745098039216);
  rgb(4pt)=(0.619607843137255,0.603921568627451,0.784313725490196);
  rgb(5pt)=(0.501960784313725,0.490196078431373,0.729411764705882);
  rgb(6pt)=(0.415686274509804,0.317647058823529,0.63921568627451);
  rgb(7pt)=(0.329411764705882,0.152941176470588,0.56078431372549);
  rgb(8pt)=(0.247058823529412,0,0.490196078431373)
},
point meta max=1034,
point meta min=0,
tick align=outside,
tick pos=left,
x grid style={white!69.0196078431373!black},
xlabel={Predicted label},
xmin=-0.5, xmax=4.5,
xtick style={color=black},
xtick={0,1,2,3,4},
xticklabels={takeoff,land,forward,cw,ccw},
y dir=reverse,
y grid style={white!69.0196078431373!black},
ylabel={True label},
ymin=-0.5, ymax=4.5,
ytick style={color=black},
ytick={0,1,2,3,4},
yticklabels={takeoff,land,forward,cw,ccw}
]
\addplot graphics [includegraphics cmd=\pgfimage,xmin=-0.5, xmax=4.5, ymin=4.5, ymax=-0.5] {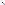};
\draw (axis cs:0,0) node[
  scale=0.789473684210526,
  text=black,
  rotate=0.0
]{175};
\draw (axis cs:1,0) node[
  scale=0.789473684210526,
  text=black,
  rotate=0.0
]{0};
\draw (axis cs:2,0) node[
  scale=0.789473684210526,
  text=black,
  rotate=0.0
]{0};
\draw (axis cs:3,0) node[
  scale=0.789473684210526,
  text=black,
  rotate=0.0
]{0};
\draw (axis cs:4,0) node[
  scale=0.789473684210526,
  text=black,
  rotate=0.0
]{0};
\draw (axis cs:0,1) node[
  scale=0.789473684210526,
  text=black,
  rotate=0.0
]{0};
\draw (axis cs:1,1) node[
  scale=0.789473684210526,
  text=black,
  rotate=0.0
]{151};
\draw (axis cs:2,1) node[
  scale=0.789473684210526,
  text=black,
  rotate=0.0
]{9};
\draw (axis cs:3,1) node[
  scale=0.789473684210526,
  text=black,
  rotate=0.0
]{7};
\draw (axis cs:4,1) node[
  scale=0.789473684210526,
  text=black,
  rotate=0.0
]{6};
\draw (axis cs:0,2) node[
  scale=0.789473684210526,
  text=black,
  rotate=0.0
]{0};
\draw (axis cs:1,2) node[
  scale=0.789473684210526,
  text=black,
  rotate=0.0
]{2};
\draw (axis cs:2,2) node[
  scale=0.789473684210526,
  text=white,
  rotate=0.0
]{1034};
\draw (axis cs:3,2) node[
  scale=0.789473684210526,
  text=black,
  rotate=0.0
]{69};
\draw (axis cs:4,2) node[
  scale=0.789473684210526,
  text=black,
  rotate=0.0
]{99};
\draw (axis cs:0,3) node[
  scale=0.789473684210526,
  text=black,
  rotate=0.0
]{0};
\draw (axis cs:1,3) node[
  scale=0.789473684210526,
  text=black,
  rotate=0.0
]{4};
\draw (axis cs:2,3) node[
  scale=0.789473684210526,
  text=black,
  rotate=0.0
]{50};
\draw (axis cs:3,3) node[
  scale=0.789473684210526,
  text=white,
  rotate=0.0
]{541};
\draw (axis cs:4,3) node[
  scale=0.789473684210526,
  text=black,
  rotate=0.0
]{95};
\draw (axis cs:0,4) node[
  scale=0.789473684210526,
  text=black,
  rotate=0.0
]{0};
\draw (axis cs:1,4) node[
  scale=0.789473684210526,
  text=black,
  rotate=0.0
]{9};
\draw (axis cs:2,4) node[
  scale=0.789473684210526,
  text=black,
  rotate=0.0
]{88};
\draw (axis cs:3,4) node[
  scale=0.789473684210526,
  text=black,
  rotate=0.0
]{62};
\draw (axis cs:4,4) node[
  scale=0.789473684210526,
  text=white,
  rotate=0.0
]{599};
\end{axis}

\end{tikzpicture}
\caption{Confusion matrix \textit{diagonal camera} model (83.3\% accuracy) }
\label{fig:conf_matrix_diagonal826}
\end{figure}
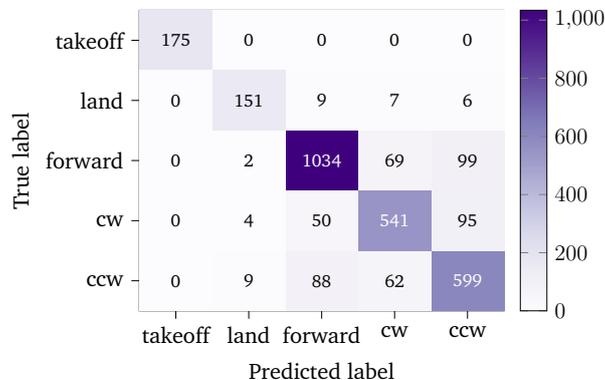

Due to the fact that the two previously discussed models had very opposing strengths and only differed in regards to the camera angle, one might think that a camera angle in between 10\textdegree{} and 90\textdegree{} combines the strengths of both models. We have therefore created a dataset using a camera angle of 45\textdegree{} (and 82.6\textdegree{} DFOV) and trained a new model (hereafter referred to as ``diagonal camera model"). Evaluating its performance on the test set revealed a prediction accuracy of 83.3\% which is significantly better than the performance of the two previous models. The confusion matrix in Figure~\ref{fig:conf_matrix_diagonal826} illustrates that the performance improvement results from the fact that the diagonal model is capable of both landing and flying towards the landing platform. 

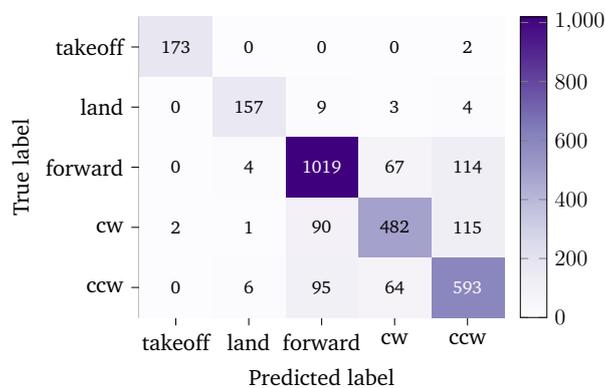
\begin{figure}[!htb]
\centering
\begin{tikzpicture}[scale=0.7]

\begin{axis}[
xlabel style={yshift=-0.3cm},
ylabel style={yshift=1cm},
colorbar,
colorbar style={ylabel={}},
colormap={mymap}{[1pt]
  rgb(0pt)=(0.988235294117647,0.984313725490196,0.992156862745098);
  rgb(1pt)=(0.937254901960784,0.929411764705882,0.96078431372549);
  rgb(2pt)=(0.854901960784314,0.854901960784314,0.92156862745098);
  rgb(3pt)=(0.737254901960784,0.741176470588235,0.862745098039216);
  rgb(4pt)=(0.619607843137255,0.603921568627451,0.784313725490196);
  rgb(5pt)=(0.501960784313725,0.490196078431373,0.729411764705882);
  rgb(6pt)=(0.415686274509804,0.317647058823529,0.63921568627451);
  rgb(7pt)=(0.329411764705882,0.152941176470588,0.56078431372549);
  rgb(8pt)=(0.247058823529412,0,0.490196078431373)
},
point meta max=1019,
point meta min=0,
tick align=outside,
tick pos=left,
x grid style={white!69.0196078431373!black},
xlabel={Predicted label},
xmin=-0.5, xmax=4.5,
xtick style={color=black},
xtick={0,1,2,3,4},
xticklabels={takeoff,land,forward,cw,ccw},
y dir=reverse,
y grid style={white!69.0196078431373!black},
ylabel={True label},
ymin=-0.5, ymax=4.5,
ytick style={color=black},
ytick={0,1,2,3,4},
yticklabels={takeoff,land,forward,cw,ccw}
]
\addplot graphics [includegraphics cmd=\pgfimage,xmin=-0.5, xmax=4.5, ymin=4.5, ymax=-0.5] {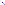};
\draw (axis cs:0,0) node[
  scale=0.789473684210526,
  text=black,
  rotate=0.0
]{173};
\draw (axis cs:1,0) node[
  scale=0.789473684210526,
  text=black,
  rotate=0.0
]{0};
\draw (axis cs:2,0) node[
  scale=0.789473684210526,
  text=black,
  rotate=0.0
]{0};
\draw (axis cs:3,0) node[
  scale=0.789473684210526,
  text=black,
  rotate=0.0
]{0};
\draw (axis cs:4,0) node[
  scale=0.789473684210526,
  text=black,
  rotate=0.0
]{2};
\draw (axis cs:0,1) node[
  scale=0.789473684210526,
  text=black,
  rotate=0.0
]{0};
\draw (axis cs:1,1) node[
  scale=0.789473684210526,
  text=black,
  rotate=0.0
]{157};
\draw (axis cs:2,1) node[
  scale=0.789473684210526,
  text=black,
  rotate=0.0
]{9};
\draw (axis cs:3,1) node[
  scale=0.789473684210526,
  text=black,
  rotate=0.0
]{3};
\draw (axis cs:4,1) node[
  scale=0.789473684210526,
  text=black,
  rotate=0.0
]{4};
\draw (axis cs:0,2) node[
  scale=0.789473684210526,
  text=black,
  rotate=0.0
]{0};
\draw (axis cs:1,2) node[
  scale=0.789473684210526,
  text=black,
  rotate=0.0
]{4};
\draw (axis cs:2,2) node[
  scale=0.789473684210526,
  text=white,
  rotate=0.0
]{1019};
\draw (axis cs:3,2) node[
  scale=0.789473684210526,
  text=black,
  rotate=0.0
]{67};
\draw (axis cs:4,2) node[
  scale=0.789473684210526,
  text=black,
  rotate=0.0
]{114};
\draw (axis cs:0,3) node[
  scale=0.789473684210526,
  text=black,
  rotate=0.0
]{2};
\draw (axis cs:1,3) node[
  scale=0.789473684210526,
  text=black,
  rotate=0.0
]{1};
\draw (axis cs:2,3) node[
  scale=0.789473684210526,
  text=black,
  rotate=0.0
]{90};
\draw (axis cs:3,3) node[
  scale=0.789473684210526,
  text=black,
  rotate=0.0
]{482};
\draw (axis cs:4,3) node[
  scale=0.789473684210526,
  text=black,
  rotate=0.0
]{115};
\draw (axis cs:0,4) node[
  scale=0.789473684210526,
  text=black,
  rotate=0.0
]{0};
\draw (axis cs:1,4) node[
  scale=0.789473684210526,
  text=black,
  rotate=0.0
]{6};
\draw (axis cs:2,4) node[
  scale=0.789473684210526,
  text=black,
  rotate=0.0
]{95};
\draw (axis cs:3,4) node[
  scale=0.789473684210526,
  text=black,
  rotate=0.0
]{64};
\draw (axis cs:4,4) node[
  scale=0.789473684210526,
  text=white,
  rotate=0.0
]{593};
\end{axis}

\end{tikzpicture}
\caption{Confusion matrix \textit{half front and half bottom camera} model (80.8\% accuracy) }
\label{fig:conf_matrix_combi826}
\end{figure}

Another attempt in combining the strengths of the front and bottom camera model was to join the two images into one. Such an image could on the real drone be created by attaching a mirror in front of the camera. This mirror splits the view of the camera in half such that the lower half shows the forward view and the upper half shows the downward view. We have again generated a dataset using this camera set-up, trained a new model (hereafter referred to as ``half front and half bottom model") which is based on the same CNN architecture as the previous models, and evaluated its performance on the test set. It achieved a prediction accuracy of 80.8\% which is slightly worse than the performance of the diagonal model. Figure~\ref{fig:conf_matrix_combi826} illustrates that the model is also capable of predicting all flight commands with high accuracy.

\begin{figure}[!htb]
\centering
\begin{tikzpicture}[scale=0.7]

\begin{axis}[
xlabel style={yshift=-0.3cm},
ylabel style={yshift=1cm},
colorbar,
colorbar style={ylabel={}},
colormap={mymap}{[1pt]
  rgb(0pt)=(0.988235294117647,0.984313725490196,0.992156862745098);
  rgb(1pt)=(0.937254901960784,0.929411764705882,0.96078431372549);
  rgb(2pt)=(0.854901960784314,0.854901960784314,0.92156862745098);
  rgb(3pt)=(0.737254901960784,0.741176470588235,0.862745098039216);
  rgb(4pt)=(0.619607843137255,0.603921568627451,0.784313725490196);
  rgb(5pt)=(0.501960784313725,0.490196078431373,0.729411764705882);
  rgb(6pt)=(0.415686274509804,0.317647058823529,0.63921568627451);
  rgb(7pt)=(0.329411764705882,0.152941176470588,0.56078431372549);
  rgb(8pt)=(0.247058823529412,0,0.490196078431373)
},
point meta max=1066,
point meta min=0,
tick align=outside,
tick pos=left,
x grid style={white!69.0196078431373!black},
xlabel={Predicted label},
xmin=-0.5, xmax=4.5,
xtick style={color=black},
xtick={0,1,2,3,4},
xticklabels={takeoff,land,forward,cw,ccw},
y dir=reverse,
y grid style={white!69.0196078431373!black},
ylabel={True label},
ymin=-0.5, ymax=4.5,
ytick style={color=black},
ytick={0,1,2,3,4},
yticklabels={takeoff,land,forward,cw,ccw}
]
\addplot graphics [includegraphics cmd=\pgfimage,xmin=-0.5, xmax=4.5, ymin=4.5, ymax=-0.5] {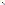};
\draw (axis cs:0,0) node[
  scale=0.789473684210526,
  text=black,
  rotate=0.0
]{175};
\draw (axis cs:1,0) node[
  scale=0.789473684210526,
  text=black,
  rotate=0.0
]{0};
\draw (axis cs:2,0) node[
  scale=0.789473684210526,
  text=black,
  rotate=0.0
]{0};
\draw (axis cs:3,0) node[
  scale=0.789473684210526,
  text=black,
  rotate=0.0
]{0};
\draw (axis cs:4,0) node[
  scale=0.789473684210526,
  text=black,
  rotate=0.0
]{0};
\draw (axis cs:0,1) node[
  scale=0.789473684210526,
  text=black,
  rotate=0.0
]{0};
\draw (axis cs:1,1) node[
  scale=0.789473684210526,
  text=black,
  rotate=0.0
]{157};
\draw (axis cs:2,1) node[
  scale=0.789473684210526,
  text=black,
  rotate=0.0
]{10};
\draw (axis cs:3,1) node[
  scale=0.789473684210526,
  text=black,
  rotate=0.0
]{4};
\draw (axis cs:4,1) node[
  scale=0.789473684210526,
  text=black,
  rotate=0.0
]{2};
\draw (axis cs:0,2) node[
  scale=0.789473684210526,
  text=black,
  rotate=0.0
]{0};
\draw (axis cs:1,2) node[
  scale=0.789473684210526,
  text=black,
  rotate=0.0
]{4};
\draw (axis cs:2,2) node[
  scale=0.789473684210526,
  text=white,
  rotate=0.0
]{1066};
\draw (axis cs:3,2) node[
  scale=0.789473684210526,
  text=black,
  rotate=0.0
]{45};
\draw (axis cs:4,2) node[
  scale=0.789473684210526,
  text=black,
  rotate=0.0
]{89};
\draw (axis cs:0,3) node[
  scale=0.789473684210526,
  text=black,
  rotate=0.0
]{0};
\draw (axis cs:1,3) node[
  scale=0.789473684210526,
  text=black,
  rotate=0.0
]{3};
\draw (axis cs:2,3) node[
  scale=0.789473684210526,
  text=black,
  rotate=0.0
]{53};
\draw (axis cs:3,3) node[
  scale=0.789473684210526,
  text=white,
  rotate=0.0
]{586};
\draw (axis cs:4,3) node[
  scale=0.789473684210526,
  text=black,
  rotate=0.0
]{48};
\draw (axis cs:0,4) node[
  scale=0.789473684210526,
  text=black,
  rotate=0.0
]{0};
\draw (axis cs:1,4) node[
  scale=0.789473684210526,
  text=black,
  rotate=0.0
]{5};
\draw (axis cs:2,4) node[
  scale=0.789473684210526,
  text=black,
  rotate=0.0
]{64};
\draw (axis cs:3,4) node[
  scale=0.789473684210526,
  text=black,
  rotate=0.0
]{36};
\draw (axis cs:4,4) node[
  scale=0.789473684210526,
  text=white,
  rotate=0.0
]{653};
\end{axis}

\end{tikzpicture}
\caption{Confusion matrix \textit{fisheye front camera} model (87.9\% accuracy) }
\label{fig:conf_matrix_front150}
\end{figure}
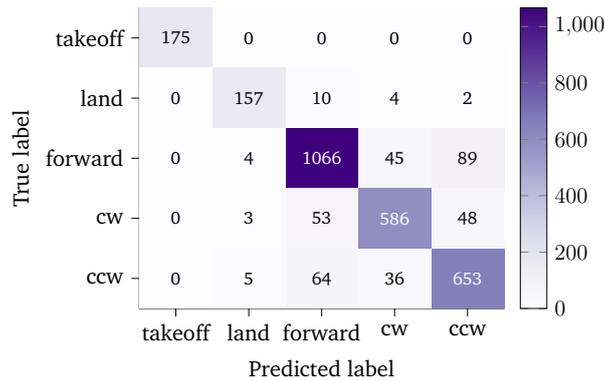

The last camera set-up that was tested uses a forward facing camera with a fisheye lens. Fisheye lenses are intended to create panoramic images with an ultra-wide DFOV. We have generated a dataset using a forward facing camera (tilted 10\textdegree{} downward) with a DFOV of 150\textdegree{}. The model that has been trained on this dataset (hereafter referred to as ``fisheye front camera model") achieves a prediction accuracy of 87.9\% which is higher than the accuracy of all other models. The confusion matrix in Figure~\ref{fig:conf_matrix_front150} illustrates that the model is capable of predicting all five flight commands with high accuracy, just like the diagonal model and the half front and half bottom model, but outperforms both in regards to the prediction of the forward and rotational commands. 

After having evaluated several camera set-ups, it was necessary to make an educated decision on which set-up to choose. For this decision, the front camera model and the bottom camera model were not considered as we have already seen that these can't predict the comprehensive set of flight commands. As depicted in Table~\ref{tab:acc_cam} the fisheye front camera model achieved the highest accuracy on the test set. But since accuracy only captures the overall performance, we needed to analyse the different types of misclassification in order to get a better picture of the model's performance. In regards to our task at hand, not all misclassifications are equally severe. For example, under the assumption that a flight is terminated after landing, misclassifying a cw command as a land command will most likely result in an unsuccessful flight because the drone landed outside the landing platform. In contrast, misclassifying a cw command as ccw might increase the flight time by a few seconds, but can still lead to a successful landing. 

Let's have a look at the different types of misclassification, beginning with the takeoff command. Misclassifying the takeoff command is severe, because in that case the drone wouldn't be able to start the flight. This type of misclassification is, however, very rare across all models. In regards to the land command, both Recall and Precision are important metrics to consider. If a flight command is falsely classified as a land command, the drone will most likely land outside the landing platform. In contrast, if the ground truth label is land, but the model predicts another flight command, this might cause the drone to fly away from the platform. Recall and Precision metrics for the land command are all at a very similar level across the three adjusted camera set-ups. Regarding the forward command, Recall is more important than Precision. This is due to the fact that a falsely predicted forward command can result in the drone crashing into an obstacle which is the worst possible outcome. When comparing the Recall of the forward command amongst the three adjusted camera set-ups, we find that the fisheye camera model has a forward command Recall of 89.35\% which is roughly two percentage points higher than the metric of the diagonal camera model and more than five percentage points higher than the metric of the half front and half bottom model. Lastly, misclassifying a rotational commands is less severe as this will most likely only prolong the flight. Nevertheless, the fisheye camera set-up also outperforms the other models in regards to the rotational commands. Due to its simplicity and all the above mentioned factors, the fisheye camera set-up has been chosen for the further course of this project.

\begin{table}[!htb]
\centering
\caption{Accuracy comparison of camera set-ups}
\begin{tabular}{c c}
\hline
\hline
\textbf{Camera Model Set-Up} & \textbf{Accuracy} \\
\hline
Front & 67.6\% \\
Bottom & 60.5\% \\
Diagonal & 83.3\% \\
Half From and Half Bottom & 80.8\% \\
Fisheye & 87.9\% \\
\hline
\hline
\end{tabular}
\label{tab:acc_cam}
\end{table}

\newpage

\section{Neural Network Architecture}

\begin{figure}[!htb]
\centering
\includegraphics[width=\textwidth]{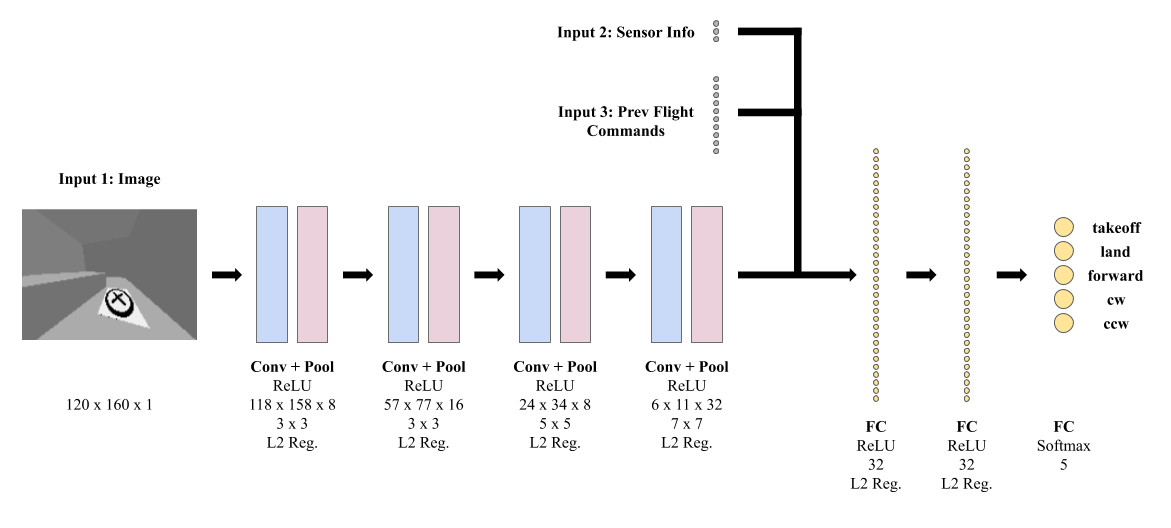}
\caption{Optimised neural network architecture}
\label{fig:cnn_opti}
\end{figure}

Now that we have determined a suitable data collection procedure and camera set-up, we can construct our neural network. The task which this neural network attempts to solve is to predict the optimal flight command which the drone should execute next. We are therefore faced with a multi-class classification problem because the neural network needs to predict one out of five potential flight commands. The inputs to the neural network are threefold: an image taken with the drone's camera, sensor information captured by the drone's sensors and the previously executed flight commands. Based on these input and output requirements we developed an optimal neural network architecture, which is depicted in Figure~\ref{fig:cnn_opti}.

This architecture was optimised based on two metrics: macro-averaged F1-score (macro F1-score) and latency. The macro F1-score is calculated by computing the F1-score (harmonic mean between precision and recall) for each class and then averaging it across all classes using the arithmetic mean \citep{opitz}:

\begin{equation}
\text{macro F1-score} = \frac{1}{n} \sum_{x} \frac{2 P_x R_x}{P_x + R_x}
\end{equation}

where:
\begin{conditions}
n & number of classes \\
P & precision \\   
R & recall
\end{conditions}

We chose the macro F1-score because our dataset is slightly imbalanced and because we regard each of the five flight commands as equally important. If we used accuracy instead, the overrepresented classes in the dataset would have a higher impact on the performance measure than the underrepresented ones. Since the macro F1-score gives each class the same weight it can better capture the true performance of the model and is thus a better metric than accuracy. 

Furthermore, latency has been chosen as a second optimisation metric, because the model needs to make predictions in real time. These predictions thus need to be quick to make the drone control more efficient. One can reduce the prediction latency of a model by using compact architectures with a fewer number of parameters, low resolution input images and low-precision number representations. 

Optimising a CNN architecture is a complex task as there is a large number of parameters which can be tuned. In order to find an optimal neural network architecture, we used the following strategy: At first, we laid out the foundations, including the dataset as well as the structure of the input and output layers. We then compared basic CNN architectures against each other and chose the best performing one. In a next step, we tested whether we can further reduce the complexity of the network by using lower resolution input images and low-precision number representations without sacrificing performance. Thereafter, we introduces techniques to avoid overfitting. And in a final optimisation step, we performed hyper parameter tuning in order to further increase the macro F1-score. The following subsections describe the optimisation procedure in greater detail. 

\subsection{Foundations}
\label{sec:foundations}

Before we can compare different architectures, we first need to clarify the dataset, the structure of the input layers and the structure of the output layer.

\subsubsection{Dataset}

The dataset, that we used for optimising the model, includes 100\,000 data samples which have been collected using the previously described data collection procedure over a total of 5\,125 flights. For each of these flights, the drone, the landing platform and a number of obstacles have been randomly placed in the simulation environment. The number of obstacles (between 0 and 10) and the edge lengths of the obstacles (between 0cm and 200cm) have been randomly chosen using a pseudo-random uniform distribution. Each flight was thus conducted in a different simulation environment. The dataset has been partitioned into a train, validation and test set using a 70/15/15 split, while samples captured during the same flight were required to be in the same partition to avoid similarities across the partitions. The label distributions for each of these datasets can be found in Appendix~\ref{app:dataset}.

\subsubsection{Input Layers}

As mentioned earlier, the model should take in an image, sensor information and the previously executed flight commands. Let's discuss each of these inputs so we get a better understanding of them. 

The image input layer takes in a gray scale image. We are using gray scale images because we want to keep the model as simple as possible and because we aren't using any colors in the simulation environment. Pixel values are normalized in a preprocessing step to a range between 0 and 1 in order to make them more suitable for the model, because high input values can disrupt the learning process. Since we have already determined that the fish-eye set-up is optimal for our purpose, the images in the dataset were taken using a forward facing camera with a DFOV of 150\textdegree{} tilted 10\textdegree{} downward. We started our optimisation procedure using an image dimension of 280 x 210 pixels.

The sensor input layer takes in three floating point values: the height relative to the takeoff position in meters, the euclidean distance relative to the takeoff position in meters and the number of flight commands executed since takeoff.

The third input layer takes in the previously executed flight commands. The intuition behind this layer is to add a sort of memory to the neural network such that it can consider historic data when making its prediction. Flight commands are categorical values which can't be interpreted directly by the neural network. Simply translating the flight commands to integer values, like we are doing for the labels of the dataset, also won't work because it doesn't make sense to rank the flight commands. Instead, we use one-hot encoding to translate a flight command into five boolean values. For example, if the previous flight command was the ``forward" command, then the boolean value for ``forward" is set to 1 and all other boolean values which resemble the remaining four flight commands are set to 0. We could apply the same logic not only to the flight command that was executed last, but also the second-last, third-last and so forth. For each additional flight command, we need to add five additional boolean values to the layer. This layout ensures that the neural network can make sense of the previously executed flight commands. We started our optimisation procedure using the two previously executed flight commands, which translate into an input layer of ten boolean values. 

\subsubsection{Output Layer}

Since we are dealing with a multi-class classification problem, the final layer is made up five neurons (one for each class) and is using the Softmax activation function. Softmax transforms real values into probabilities in a range [0,1] which sum up to 1. The model thus outputs a probability for each class and using the argmax function we can determine the index of the class with the highest probability. This class corresponds to the flight command which the model has predicted. 

The default loss function for a multi-class classification problem is categorical cross-entropy loss. This loss function evaluates how well the probability distribution produced by the model matches the one-hot encoded label of the data sample. Since our dataset is balanced according to how often each command is on average executed during flight, it makes sense to add class weights to the loss functions. These class weights need to be higher for under represented classes (i.e. ``takeoff" and ``land") such that the loss function penalises the model more if one of these classes is misclassified. Table~\ref{tab:weights} shows the weight for each class that is used for the cross-entropy loss function. The weights have been calculated by dividing the share of the majority class by the share of each class. 

\begin{table}[!htb]
\centering
\caption{Class weights used for loss function}
\begin{tabular}{c c c}
\hline
\hline
\textbf{Flight Command} & \textbf{Relative Share} & \textbf{Weight} \\
\hline
takeoff & ~5\% & 8 \\
land & ~5\% & 8 \\
forward & ~40\% & 1 \\
cw & ~25\% & 1.6 \\
ccw & ~25\% & 1.6 \\
\hline
\hline
\end{tabular}
\label{tab:weights}
\end{table}

\subsection{Architecture Comparison}

The state of the art neural network architectures for image processing are CNNs. CNNs typically begin with convolutional and pooling layers which extract information from the image by interpreting pixel values in relation to their neighbouring pixels. The deeper you go in these layers, the more high level pixel relationships can be detected. Convolutional and pooling layers are then followed by fully connected layers which classify the image. Since it doesn't make sense to convolve sensor information or the previously executed flight commands, we can feed these two input layers directly into the first fully connected layer.

Now that we have clarified the overall structure of the model, we need to find a combination of convolutional, pooling and fully connected layers which work best for our task. We created eight different architectures which have been inspired by LeNet, AlexNet and VGG-16 \citep{lenet, alexnet, vgg}. These architectures only differ in the number and order of convolutional, pooling and fully connected layers.  All other parameters, such as learning rate, epochs, batch size, number of filters per convolution, stride, padding, kernel size, number of neurons per fully connected layer and activation functions, are the same across all eight architectures. The values which have been used for these parameters can be found in Appendix~\ref{app:arch_comp}.

\begin{table}[!htb]
\centering
\caption{Architecture comparison}
\begin{threeparttable}
\begin{tabular}{c l c c c}
\hline
\hline
\textbf{ID} & \textbf{Architecture}\tnote{1} & \textbf{\# Parameters} & \textbf{Macro F1-score}\tnote{2} \\
\hline
1 & C-P-C-P-F & 3'561'413 & 93.01\% (3) \\
2 & C-P-C-P-F-F & 3'562'469 & 93.09\% (3) \\
\\
3 & C-P-C-P-C-P-F & 830'437 & 92.58\% (5) \\
4 & C-P-C-P-C-P-F-F & 831'493 & 93.04\% (6) \\
\\
5 & C-P-C-P-C-P-C-P-F & 197'637 & 92.97\% (9) \\
6 & C-P-C-P-C-P-C-P-F-F & 198'693 & 93.18\% (6) \\
\\
7 & C-C-P-C-C-P-F & 3'390'469 & 92.71\% (2) \\
8 & C-C-P-C-C-P-F-F & 3'391'525 & 93.16\% (3) \\
\hline
\hline
\end{tabular}
\begin{tablenotes}
\item[1] C = convolutional layer, P = pooling layer, F = fully connected layer
\item[2] Highest macro F1-score on validation set across ten epochs. Epoch given in brackets.
\end{tablenotes}
\end{threeparttable}
\label{tab:model_comp_archi}
\end{table}

For each architecture, we trained a model on the train set for ten epochs and evaluated its performance on the validation set after every epoch. Table~\ref{tab:model_comp_archi} shows the number of trainable parameters for each model as well as the highest accuracy and the highest macro F1-score that have been achieved on the validation set. The numbers in brackets represent the epochs with the highest measurements. There are four main takeaways from this table: Firstly, we can see that all architectures achieve high macro F1-scores which vary only within a range of 0.6 percentage points. Nevertheless, Architectures 6 and 8 have the highest macro F1-scores. Secondly, we can see that the number of parameters significantly varies across the architectures, with Architectures 5 and 6 being the smallest. Thirdly, we can see that models with a higher number of parameters are very prone to overfitting since they reach their highest macro F1-scores on the validation set after only a few number of epochs. 
And fourthly, we can see that models with two fully connected layers are superior to models with one fully connected layer. 

Since we are optimising the model on macro F1-score and latency, we are interested in a model with a high macro F1-score and a low number of parameters (since this reduces the latency). Considering the main takeaways from Table~\ref{tab:model_comp_archi}, we can conclude that Architecture 6 is the optimal choice because it not only has the highest macro F1-score but also the second lowest number of parameters.

\subsection{Input Layer Tuning}

The next step in optimising the neural network concerns the input layers. Two of the three input layers provide room for optimisation. Regarding the image input layer, we can change the resolution of the images. And concerning the previously executed flight command input layer, we can change how many flight commands are provided to the network. 

\begin{figure}[!htb]
\centering
\begin{tikzpicture}[scale=0.8,
declare function={
    barW=20pt; 
    barShift=barW/2; 
  }]
\begin{axis}
[
width=16cm,
height=8cm,
xlabel={Image Resolution in Pixels}, 
xlabel style={yshift=-1.4cm},
xtick=data,
xticklabel style={rotate=45},
symbolic x coords={80 x 60, 120 x 90, 160 x 120, 200 x 150, 240 x 180, 280 x 210, 320 x 240},
ylabel={Macro F1},
ylabel style={yshift=0cm, color = blue},
yticklabel={$\pgfmathprintnumber{\tick}\%$},
ytick pos=left,
ymin=89, 
ymax=94, 
enlarge x limits=0.15,
ybar,
bar width=barW, 
bar shift=-barShift, 
nodes near coords align={vertical}, 
nodes near coords={\pgfmathprintnumber\pgfplotspointmeta\%},
]
\addplot coordinates {(80 x 60,90.64) (120 x 90,91.65) (160 x 120,93.18) (200 x 150,92.30) (240 x 180,92.83) (280 x 210,93.18) (320 x 240,92.86) };
\end{axis}

\begin{axis}
[
width=16cm,
height=8cm,
axis y line*=right,
ylabel={\# Parameters},
ylabel style={yshift=-17.1cm, color = red},
yticklabel={$\pgfmathprintnumber{\tick}$k},
ymin=0, 
ymax=300, 
axis x line=none,
symbolic x coords={60 x 80, 90 x 120, 120 x 160, 150 x 200, 180 x 240, 210 x 280, 240 x 320},
enlarge x limits=0.15,
ybar,
bar width=barW,
bar shift=barShift,
]
\definecolor{r1}{RGB}{255, 178, 178}
\definecolor{r2}{RGB}{255, 0, 0}
\addplot+[mark=x, draw=r2, fill=r1] coordinates {(60 x 80,33) (90 x 120,45) (120 x 160,71) (150 x 200,101) (180 x 240,150) (210 x 280, 199) (240 x 320, 269) };
\end{axis}

\end{tikzpicture}
\caption{Performance evaluation with varying image resolutions}
\label{fig:resolution}
\end{figure}
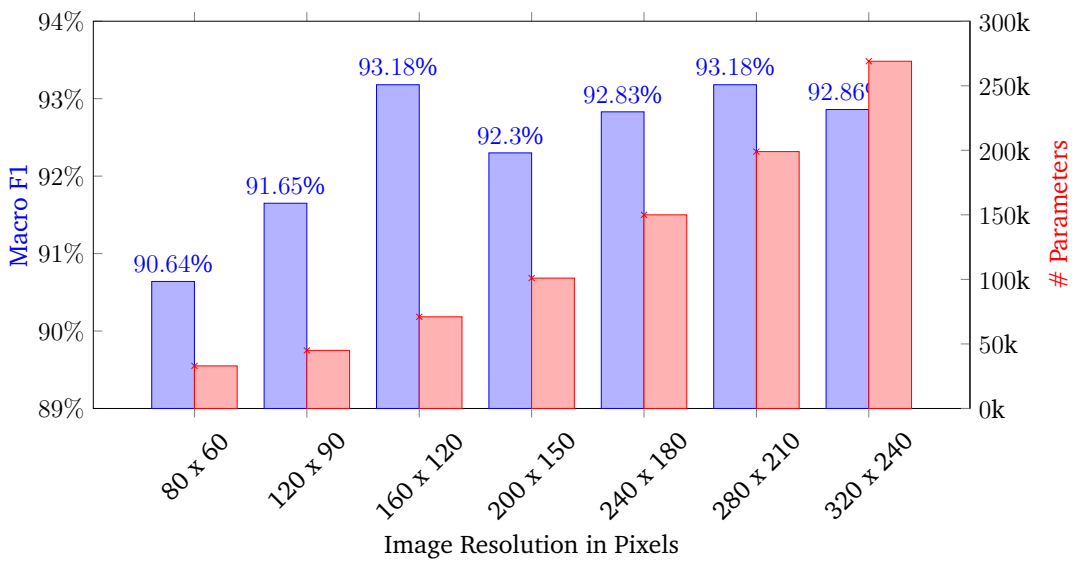

Let's begin with optimising the image input layer. We have trained seven different models that are all based on Architecture 6, but vary in the resolution of input images. All models have been trained for ten epochs on the same train set (with adjusted image resolutions) and validated after each epoch on the validation set. Figure~\ref{fig:resolution} provides the highest macro F1-score across the ten epochs as well as the number of parameters for each of the seven models. The red line illustrates that smaller image resolutions reduce the complexity of the model which is again beneficial for the prediction latency. Reducing the resolution of the image, however, also means that the model has fewer inputs on which it can base its prediction, resulting in a lower macro F1-score. This relationship is illustrated by the blue line in Figure~\ref{fig:resolution}, but we can see a spike regarding the macro F1-score for the 160 x 120 image resolution. This macro F1-score is exactly the same as the macro F1-score of the model with an image resolution of 280 x 210, but considering a reduction of 64\% in the number of parameters, the model with an image resolution of 160 x 120 is the superior choice. 

\begin{figure}[!htb]
\centering
\begin{tikzpicture}[scale=0.8]
\begin{axis}
[
width=12cm,
height=8cm,
xlabel={\# Prev. Flight Command Input}, 
xlabel style={yshift=-0.3cm},
xtick=data,
xticklabel style={rotate=0},
symbolic x coords={0, 1, 2, 3, 4, 5},
ylabel={Macro F1},
ylabel style={yshift=0cm},
yticklabel={$\pgfmathprintnumber{\tick}\%$},
ymin=89, 
ymax=94, 
enlarge x limits=0.15,
nodes near coords align={vertical}, 
nodes near coords={\pgfmathprintnumber\pgfplotspointmeta\%},
]
\addplot coordinates {(0,89.28) (1,92.99) (2,93.18) (3,92.97) (4,92.95) (5,92.80)};
\end{axis}
\end{tikzpicture}
\caption{Performance evaluation with varying number of previously executed flight commands as input}
\label{fig:prev_layer}
\end{figure}
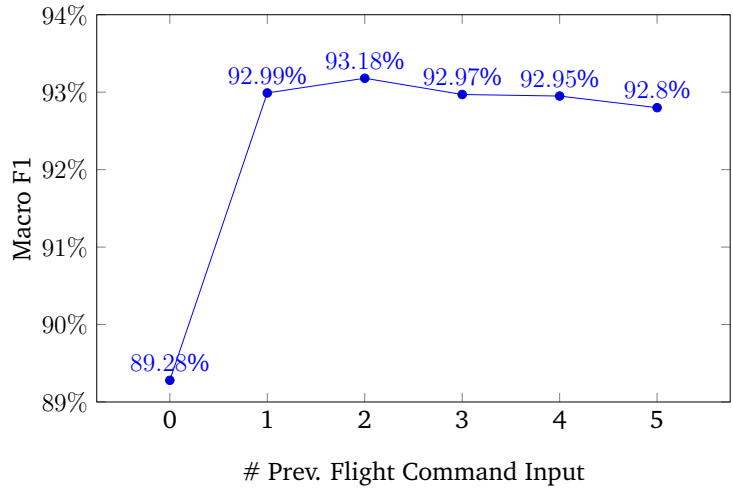

Now that we have determined the optimal image resolution, let's have a look at the input layer which captures the previously executed flight commands. We have again trained six models for ten epochs and evaluated their performance on the validation set after each period. All models are based on Architecture 6 and use an image resolution of 160 x 120, but vary in the number of previously executed flight commands that are taken as input. Remember that these flight commands are one-hot encoded, so an input layer that takes in four flight commands consists of $4*5=20$ parameters. Figure~\ref{fig:prev_layer} shows the highest macro F1-score of each model across the ten epochs. We can clearly see that the models which receive information about the previously executed flight commands perform significantly better than the model that only uses image and sensor data as input. More specifically, having knowledge about at least one previously executed flight command, increases the macro F1-score by over 2.5 percentage points. We can also see that having too many inputs decreases the performance which might be due to overfitting. Since the macro F1-score is peaking at two flight commands, we have chosen this number as the optimum for the input layer. 

One last option for tuning the input parameters was to test whether feeding the sensor and flight command input layers into its own fully connected layer (before joining it with the convolution output) increases the performance. Using this architecture we achieved a macro F1-score of 92.97\% which is worse than without an additional fully connected layer. We thus decided not to add this layer to the architecture. 

\subsection{Low-Precision Number Representations}

A further way of reducing the latency of the network is by using lower precision number representations for its weights. This correlation is due to the fact that smaller number representations consume less memory and arithmetic operations can be executed more efficiently. 

\begin{figure}[!htb]
\centering
\begin{tikzpicture}[scale=0.8]
\begin{axis}
[
width=10cm,
height=8cm,
xlabel={Floating Point Precision}, 
xlabel style={yshift=-0.3cm},
xtick=data,
xticklabel style={rotate=0},
symbolic x coords={16-bit, 16/32-bit, 32-bit, 64-bit},
ylabel={Macro F1},
ylabel style={yshift=0cm},
yticklabel={$\pgfmathprintnumber{\tick}\%$},
ymin=89, 
ymax=94, 
enlarge x limits=0.15,
nodes near coords align={vertical}, 
nodes near coords={\pgfmathprintnumber\pgfplotspointmeta\%},
ybar,
bar width=10
]
\addplot coordinates {(16-bit,92.85) (16/32-bit,93.35) (32-bit,93.18) (64-bit,92.90)};
\end{axis}
\end{tikzpicture}
\caption{Performance evaluation with varying precision number representations}
\label{fig:float}
\end{figure}
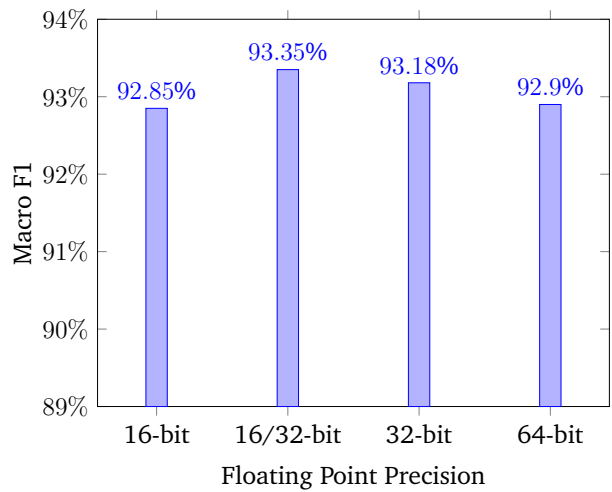

We have thus far been using 32-bit floats, but TensorFlow provides the option to change this to 16-bit, 64-bit or a mixture of 16-bit and 32-bit floats \citep{tensorflow}. Using our current architecture, we have again trained four different models with varying floating point number precisions and evaluated their performance on the validation set. Figure~\ref{fig:float} illustrates that the highest macro F1-score is achieved when using a mix between 16-bit and 32-bit precision numbers. The model with a precision of 16-bit performs the worst which can be attributed to numeric stability issues. Since the mixed precision model has both, a higher macro F1-score and a lower memory usage, compared to the 32-bit and 64-bit models, the mixed precision model has been chosen as the optimal option.

\subsection{Overfitting}

We have tested several techniques to avoid our model from overfitting. 

Our first approach was to add dropout layers after the two fully connected layers. We have trained and evaluated several models with varying dropout rates. The results showed that whilst dropout layers reduced overfitting they also negatively impacted the macro F1-score on the validation set. Based on this outcome we decided against the use of dropout layers.

In a second step, we tested L1 and L2 regularization techniques \citep{l1}. L1 regularization adds the absolute weights to the loss function and hereby encourages the model to only keep the most important features \citep{l1}. L2 regularisation, on the other hand, adds the squared weights to the loss function which pushes the weights closer to zero by penalising larger weights more and thus encourages the sharing between features \citep{l1}. We have again tested several models using either regularization techniques with a variation of $\lambda$ values. Our analysis has shown that L2 regularization with a $\lambda$ of 0.005 manages to both reduce overfitting whilst achieving a high macro F1-score on the validation set.

L2 regularization can however not prevent the model from overfitting when we are training it for too many epochs. We therefore also use an early stopping technique which aborts the training cycle when the macro F1-score on the validation set hasn't improved for several epochs.

\subsection{Hyper Parameter Tuning}

The final step in optimising the model was to perform hyper parameter tuning. We have split this up into two parts, beginning with tuning the convolutional and fully connected layers, followed by selecting an optimizer and an activation function. 

\subsubsection{Filters, Kernels and Neurons}

Regarding the convolutional layers, we decided to tune the number of filters as well as the kernel sizes. And concerning the fully connected layers, we decided to tune the number of neurons. The problem with hyper parameter tuning for CNNs is that there is a large number of possible parameter combinations. All of these parameter combinations can't be evaluated in a reasonable amount of time. In order to overcome this challenge we used two techniques. Firstly, we limited the number of possible values for each hyper parameter to a subset of values which can be found in Table~\ref{tab:subset}.

\begin{table}[!htb]
\centering
\caption{Subsets of hyper parameter values}
\begin{tabular}{c c}
\hline
\hline
\textbf{Hyper Parameter} & \textbf{Subset of Values}  \\
\hline
\# Filters & [8, 16, 32, 64] \\
Kernel Size & [3, 5, 7, 9] \\
\# Neurons & [8, 16, 32, 64] \\
\hline
\hline
\end{tabular}
\label{tab:subset}
\end{table}

But even with the limited number of hyper parameter values, there are still 1\,048\,576 possible hyper parameter combinations. We thus used random search as a second technique to overcome this problem. We trained and evaluated 530 hyper parameter combinations which were randomly selected from the subsets. The top five parameter combinations, with the highest macro F1-score on the validation set, are listed in Table~\ref{tab:random_search}. Here we can see that we managed to further improve the macro F1-score by 0.2 percentage points compared to our earlier model. We can also make out a trend amongst the top five performing models. The models have in common that their kernel sizes are small for the first convolutional layer and get larger for the convolutional layers that follow. This makes sense, because the deeper a convolutional layer is located in the network architecture, the more high level features can be detected. And in order to detect these high level features we need to consider a larger neighbourhood around the pixels.

\begin{table}[!htb]
\centering
\caption{Top five best performing hyper parameter combinations found using random search}
\begin{tabular}{c c c c c c}
\hline
\hline
\textbf{ID} & \textbf{\# Filters} & \textbf{Kernel Size} & \textbf{\# Neurons}  & \textbf{\# Paramters} & \textbf{Macro F1-score} \\
\hline
1 & (8, 16, 16, 32) & (5, 7, 3, 9) & (32, 16) & 59\,573 & 93.56\% \\
2 & (8, 16, 8, 32) & (3, 3, 5, 7) & (32, 32) & 34\,061 & 93.55\% \\
3 & (8, 32, 32, 32) & (5, 5, 3, 7) & (32, 16) & 82\,517 & 93.53\% \\
4 & (64, 8, 64, 64) & (3, 7, 3, 5) & (32, 32) & 183\,683 & 93.51\% \\
5 & (32, 8, 32, 8) & (5, 7, 7, 9) & (32, 32) & 48\,381 & 93.47\% \\
\hline
\hline
\end{tabular}
\label{tab:random_search}
\end{table}

Out of the parameter combinations listed in Table~\ref{tab:random_search}, we have selected the parameter combination with ID 2 as the optimal choice. This parameter combination achieves the second highest macro F1-score and has the smallest number of parameters.

\subsubsection{Optimizer and Activation Function}

In a last step we compared four different optimizers (Adam, Stochastic Gradient Descend (SGD), Adagrad, RMSprop) and three different activation functions (sigmoid, ReLU, Leaky ReLU) against each other. We trained twelve different models that are based on our previously determined optimal architecture and allocated each with a different combination of optimizer and activation functions. Since the optimizers are highly dependent on their learning rates, we used the optimizer specific default learning rates that are provided by TensorFlow for our comparison. The results in Table~\ref{tab:opti} show that in regards to the activation functions, Leaky ReLU and ReLU both perform much better than Sigmoid. Regarding the optimizers, Adam and RMSprop both outperform SGD and Adagrad. Based on this analysis we have decided to keep the combination of Adam optimizer and ReLU activation functions which we have used throughout this chapter. We have also tested the Adam optimizer with a variation of learning rates but found the default learning rate of 0.001 to be optimal.

\begin{table}[!htb]
\centering
\caption{Matrix of macro F1-scores using different optimizer and activation function combinations}
\begin{tabular}{c c c c c}
\hline
\hline
& \textbf{Adam} & \textbf{SGD} & \textbf{Adagrad} & \textbf{RMSprop}   \\
\textbf{Sigmoid} & 63.88\% & 90.11\% & 8.19\% & 67.99\% \\
\textbf{ReLU}  & 93.55\% & 93.13\% & 91.00\% & 93.23\% \\
\textbf{Leaky ReLU} & 93.54\% & 93.28\% & 90.47\% & 93.51\% \\
\hline
\hline
\end{tabular}
\label{tab:opti}
\end{table}

\subsection{Evaluation on Test Set}

Now that we have optimised the architectural design of the CNN, we can evaluate its performance on the test set. The test set has been set aside throughout the whole optimisation procedure and thus allows us to test how well the model performs on unseen data. We trained the final model on the train set using a batch size of 32, validated its performance after every epoch on the validation set and saved the weights. The training process was terminated once the macro F1-score on the validation set didn't improve for more than ten epochs and the weights which achieved the highest macro F1-score were selected for the final model. 

\begin{figure}[!htb]
\centering
\begin{tikzpicture}[scale=0.7]

\begin{axis}[
xlabel style={yshift=-0.3cm},
ylabel style={yshift=1cm},
colorbar,
colorbar style={ylabel={}},
colormap={mymap}{[1pt]
  rgb(0pt)=(0.988235294117647,0.984313725490196,0.992156862745098);
  rgb(1pt)=(0.937254901960784,0.929411764705882,0.96078431372549);
  rgb(2pt)=(0.854901960784314,0.854901960784314,0.92156862745098);
  rgb(3pt)=(0.737254901960784,0.741176470588235,0.862745098039216);
  rgb(4pt)=(0.619607843137255,0.603921568627451,0.784313725490196);
  rgb(5pt)=(0.501960784313725,0.490196078431373,0.729411764705882);
  rgb(6pt)=(0.415686274509804,0.317647058823529,0.63921568627451);
  rgb(7pt)=(0.329411764705882,0.152941176470588,0.56078431372549);
  rgb(8pt)=(0.247058823529412,0,0.490196078431373)
},
point meta max=5166,
point meta min=0,
tick align=outside,
tick pos=left,
x grid style={white!69.0196078431373!black},
xlabel={Predicted label},
xmin=-0.5, xmax=4.5,
xtick style={color=black},
xtick={0,1,2,3,4},
xticklabels={takeoff,land,forward,cw,ccw},
y dir=reverse,
y grid style={white!69.0196078431373!black},
ylabel={True label},
ymin=-0.5, ymax=4.5,
ytick style={color=black},
ytick={0,1,2,3,4},
yticklabels={takeoff,land,forward,cw,ccw}
]
\addplot graphics [includegraphics cmd=\pgfimage,xmin=-0.5, xmax=4.5, ymin=4.5, ymax=-0.5] {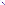};
\draw (axis cs:0,0) node[
  scale=0.789473684210526,
  text=black,
  rotate=0.0
]{766};
\draw (axis cs:1,0) node[
  scale=0.789473684210526,
  text=black,
  rotate=0.0
]{0};
\draw (axis cs:2,0) node[
  scale=0.789473684210526,
  text=black,
  rotate=0.0
]{1};
\draw (axis cs:3,0) node[
  scale=0.789473684210526,
  text=black,
  rotate=0.0
]{0};
\draw (axis cs:4,0) node[
  scale=0.789473684210526,
  text=black,
  rotate=0.0
]{1};
\draw (axis cs:0,1) node[
  scale=0.789473684210526,
  text=black,
  rotate=0.0
]{0};
\draw (axis cs:1,1) node[
  scale=0.789473684210526,
  text=black,
  rotate=0.0
]{721};
\draw (axis cs:2,1) node[
  scale=0.789473684210526,
  text=black,
  rotate=0.0
]{23};
\draw (axis cs:3,1) node[
  scale=0.789473684210526,
  text=black,
  rotate=0.0
]{7};
\draw (axis cs:4,1) node[
  scale=0.789473684210526,
  text=black,
  rotate=0.0
]{4};
\draw (axis cs:0,2) node[
  scale=0.789473684210526,
  text=black,
  rotate=0.0
]{0};
\draw (axis cs:1,2) node[
  scale=0.789473684210526,
  text=black,
  rotate=0.0
]{37};
\draw (axis cs:2,2) node[
  scale=0.789473684210526,
  text=white,
  rotate=0.0
]{5166};
\draw (axis cs:3,2) node[
  scale=0.789473684210526,
  text=black,
  rotate=0.0
]{270};
\draw (axis cs:4,2) node[
  scale=0.789473684210526,
  text=black,
  rotate=0.0
]{329};
\draw (axis cs:0,3) node[
  scale=0.789473684210526,
  text=black,
  rotate=0.0
]{0};
\draw (axis cs:1,3) node[
  scale=0.789473684210526,
  text=black,
  rotate=0.0
]{12};
\draw (axis cs:2,3) node[
  scale=0.789473684210526,
  text=black,
  rotate=0.0
]{196};
\draw (axis cs:3,3) node[
  scale=0.789473684210526,
  text=white,
  rotate=0.0
]{3603};
\draw (axis cs:4,3) node[
  scale=0.789473684210526,
  text=black,
  rotate=0.0
]{104};
\draw (axis cs:0,4) node[
  scale=0.789473684210526,
  text=black,
  rotate=0.0
]{0};
\draw (axis cs:1,4) node[
  scale=0.789473684210526,
  text=black,
  rotate=0.0
]{3};
\draw (axis cs:2,4) node[
  scale=0.789473684210526,
  text=black,
  rotate=0.0
]{205};
\draw (axis cs:3,4) node[
  scale=0.789473684210526,
  text=black,
  rotate=0.0
]{41};
\draw (axis cs:4,4) node[
  scale=0.789473684210526,
  text=white,
  rotate=0.0
]{3511};
\end{axis}

\end{tikzpicture}
\caption{Confusion matrix for test set evaluation}
\label{fig:conf_matrix_final}
\end{figure}
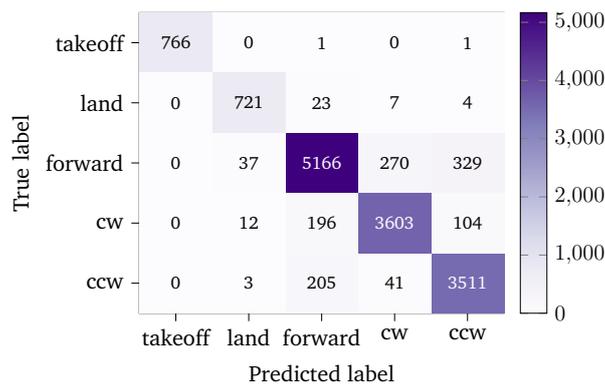

The evaluation on the test set achieved a macro F1-score of 93.60\% which is even higher than the macro F1-score on the validation set. The confusion matrix in Figure~\ref{fig:conf_matrix_final} in combination with the class-based metrics in Table~\ref{tab:class_metrics} reveal some further insights. We can see that the model is capable of predicting all five flight commands, but the performance does vary across them. Let's take a closer look at the individual flight commands. The model correctly predicts the takeoff command in all but two instances. The corresponding F1-score of 99.87\% ratifies that the model performs very well in regards to the takeoff command. The land command has the second highest F1-score across the five classes. This suggests that the drone will  likely be able to descend once it is hovering above the landing platform. The confusion matrix further illustrates that the model sometimes confuses the land with the forward command. This is understandable, since it is difficult to determine if the drone is already close enough to the landing platform in order to execute the land command. The forward command, on the other hand, is most often confused with the rotation commands. This might be caused by the fact that the landing platform can be hidden by an obstacle. In those cases, the drone can't simply rotate in order to find the platform, but needs to fly around the obstacle which requires the execution of forward commands. This explains the fact that the forward, cw and ccw flight commands have lower F1-scores than the other two flight commands.

\begin{table}[!htb]
\centering
\caption{Evaluation metrics for each class}
\begin{tabular}{c c c c}
\hline
\hline
\textbf{Flight Command} & \textbf{Precision} & \textbf{Recall} & \textbf{F1-score}  \\
\hline
takeoff & 99.74\% & 100.00\% & 99.87\% \\
land & 95.50\% & 93.27\% & 94.37\% \\
forward & 89.04\% & 92.40\% & 90.69\% \\
cw & 92.03\% & 91.89\% & 91.69\% \\
ccw & 93.38\% & 88.91\% & 91.09\% \\
\hline
\hline
\end{tabular}
\label{tab:class_metrics}
\end{table}

\section{Summary}

This chapter detailed a coherent approach regarding the development of a machine learning model for autonomous drone navigation. The approach included three major parts. 

At first, we focused on the creation of a dataset which can be used for Supervised Learning. In order to generate a dataset, we introduced a data collection procedure that is capable of producing a dataset which is balanced according to how often each flight command is executed on average during flight. The data was collected in the simulation environment, whereby the state of the drone corresponds to the features, and the optimal flight command corresponds to the label. The optimal flight commands were calculated using a BFS algorithm and we managed to significantly reduce the time complexity of the BFS algorithm by introducing several optimisation techniques. 

In a second step, we focused on finding an optimal camera set-up. We discovered that the original camera set-up of the DJI Tello drone is not suitable for the task at hand since the landing platform is getting out of sight once the drone gets to close to it. We consequently tested several other camera set-ups and identified the fisheye lens set-up as the best alternative. 

In a third step, we focused on finding an optimal CNN architecture. We achieved this by following a multistage optimisation strategy which intended to increase the macro F1-score and reduce the latency of the model. The optimised model achieved a macro F1-score of 93.60\% on the test set.

\chapter{Evaluation}
\label{chr:evaluation}

Let's recapitulate what we have done so far. In a first step, we created a simulation environment for the DJI Tello drone. In a second step, we developed a machine learning model which can be used for predicting flight commands based on visual and sensor information captured by the drone. The only thing left to do, is to put the model into action and test whether it manages to successfully navigate the drone to the landing platform. 

In this chapter, we will elaborate on the test results and qualitatively compare our approach to previous research. We have decided to solely evaluate the model in the simulation environment, and not in the real world, due to restrictions impost by the COVID-19 pandemic. 

\section{Test Flights in Simulation}
\label{sec:test_flights}

\begin{figure}[!htb]
        \centering
        \includegraphics[width=\linewidth]{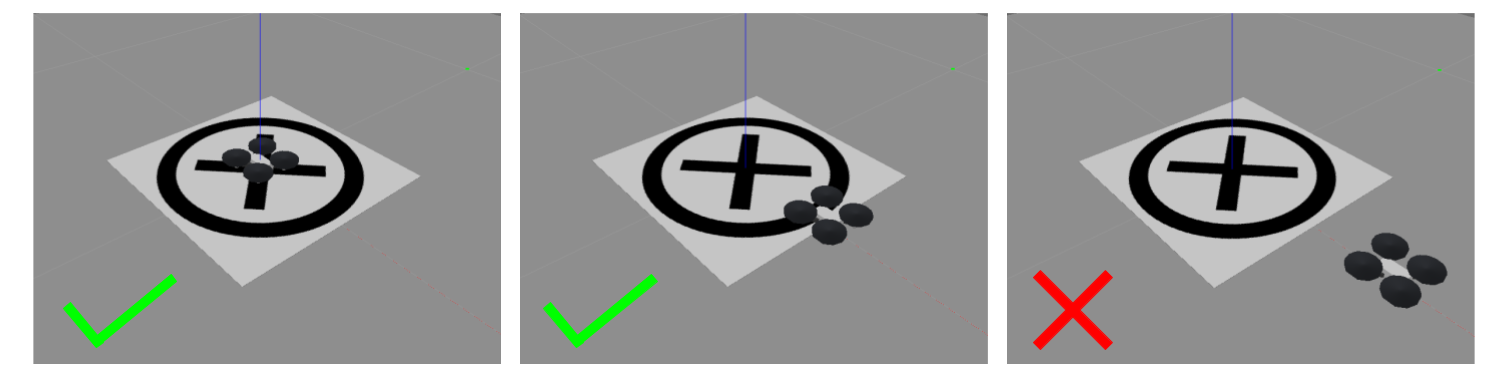}
        \caption{Examples of successful and unsuccessful landings}
        \label{fig:land}
\end{figure}

In order to evaluate the model's performance, we executed a total of 1\,000 test flights in the simulation environment. The flights were conducted in the following way: In a first step, we randomly placed the drone, the landing platform and a number of obstacles in the simulation environment. The number of obstacles (between 0 and 10) and the edge lengths of the obstacles (between 0cm and 200cm) have been randomly chosen using a pseudo-random uniform distribution. In a second step, we ensured that a potential flight path between the drone's starting position and the position of the landing platform exists. If this was not the case, which can happen if the drone is locked in by the obstacles, we generated a new simulation environment. In a third step, we let the machine learning model take over the control of the drone. This is an iterative process in which the drone continuously captures visual and sensor information, predicts a flight command based on this information and executes the predicted flight command. The flight terminates once the drone has landed somewhere, has crashed or as a default after 100 flight command executions. For each test flight, we captured several statistics such as whether drone has landed on the platform, the distance between the drones final position and the landing platform, the number of executed flight commands and the number of obstacles in the simulation environment. A test flight is classified as ``Landed on Platform" if the drone is within 30cm of the center of the landing platform as illustrated in Figure~\ref{fig:land}.

In the following, we will analyse the outcome of these test flights. We begin our analysis by looking at whether the model manages to successfully navigate the drone to the landing platform. And in a second step, we will investigate whether the flight paths that the drone has taken are efficient, by comparing them to the optimal flight paths. 

\subsection{Successful Landing Analysis}

For our first set of test flights, we trained our model on a dataset of 100\,000 data samples (using a 70\% train and 30\% validation split) and kept the model with the highest macro F1-score measured on the validation set. Using this model, the drone managed to land on the landing platform in 852 out of 1\,000 test flights. Out of the remaining 148 test flights which didn't make it onto the landing platform, the drone crashed in 141 cases, landed outside the landing platform in five cases and didn't land at all in two cases. These outcomes prove that the model is capable of successfully navigating the drone from a random starting position towards the landing platform in the majority of simulation environments. 

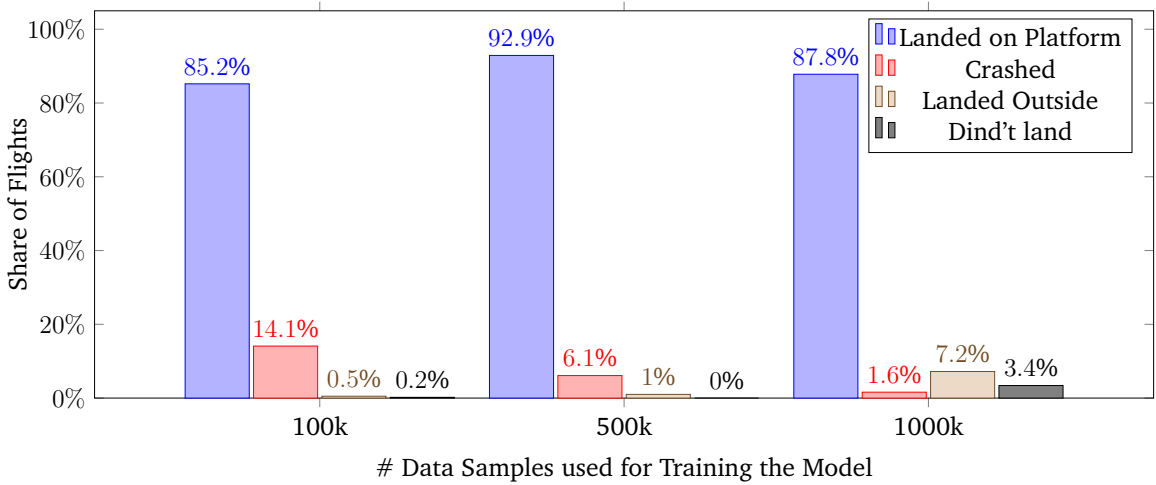
\begin{figure}[!htb]
\centering
\begin{tikzpicture}[scale=0.8]
\begin{axis}
[
width=19cm,
height=8cm,
xlabel={\# Data Samples used for Training the Model}, 
xlabel style={yshift=-0.3cm},
xtick=data,
xticklabel style={rotate=0},
symbolic x coords={100k, 500k, 1000k},
ylabel={Share of Flights},
ylabel style={yshift=0cm},
yticklabel={$\pgfmathprintnumber{\tick}\%$},
ymin=0, 
ymax=105, 
enlarge x limits=0.15,
ybar,
bar width=30,
nodes near coords align={vertical}, 
nodes near coords={\pgfmathprintnumber\pgfplotspointmeta\%},
enlarge x limits=0.37,
]
\addplot coordinates {(100k,85.2) (500k,92.9) (1000k,87.8)};
\addplot coordinates {(100k,14.1) (500k,6.1) (1000k,1.6)};
\addplot coordinates {(100k,0.5) (500k,1.0) (1000k,7.2)};
\addplot coordinates {(100k,0.2) (500k,0.0) (1000k,3.4)};
\legend{Landed on Platform, Crashed, Landed Outside, Dind't land}
\end{axis}
\end{tikzpicture}
\caption{Test flight outcomes depending on size of dataset used for training the model (70\% train and 30\% validation split)}
\label{fig:test_flight}
\end{figure}

In order to investigate whether a larger training set improves the performance, we trained two further models on a dataset of 500\,000 and 1\,000\,000 data samples, respectively, again using a 70\% train and 30\% validation split. We deployed the two models in the same 1\,000 simulation environments and under the same conditions (starting pose of drone, termination period, etc.) as the ones used for conducting the previous test flights. The outcomes of these test flights are also depicted in Figure~\ref{fig:test_flight} and can be compared against each other. We can see that the models that were trained on the larger datasets managed to navigate the drone towards the landing platform more often. This improvement is, however, not consistent since the share of successful flights is higher for the 500\,000 model than for the 1\,000\,000 model. Nevertheless, a consistent improvement can be found in regards to crashes. We can see that a larger training dataset significantly reduces the risk of crashing the drone. Depending on the use case, having a lower crash rate might even be considered to be of higher importance than making it to the landing platform more often. The model which has been trained on the dataset of 1\,000\,000 data samples is used for the proceeding analysis.

In a next step, we investigated the factors that impact whether a test flight is successful or not. For this, we have conducted the following multi linear regression:

\begin{equation}
\begin{aligned}
\mathrm{LandedOnPlatform}_i = \beta_{0} &+ \beta_{1} \mathrm{NrCuboids}_i \\
&+ \beta_{2} \mathrm{TotalCuboidVolume}_i \\
&+ \beta_{3} \mathrm{DistanceStart}_i +   \epsilon_i
\end{aligned}
\end{equation}

where:
\begin{conditions}
\beta_{0} & 1.3329 \\
\beta_{1} & -0.0315 \\
\beta_{2} & 0.0023 \\
\beta_{3} & -0.2101 \\
\end{conditions}

$\mathrm{LandedOnPlatform}$ is a boolean value which is 1 if the drone landed on the landing and 0 otherwise. The results of this regression model can be found in Appendix~\ref{app:regression}. The R-squared reveals that the model can explain 29\% of the dependent variable variation. Furthermore, we can see that the two independent variables $\mathrm{NrCuboids}$ and $\mathrm{DistanceStart}$ are highly statistically significant with a P-value \textless{} 0.001. Looking at the coefficient of to the $\mathrm{NrCuboids}$ variable, we can infer that an additional cuboid in the simulation environment reduces the chances of making it to the landing platform by 3.15 percentage points. The coefficient of the $\mathrm{DistanceStart}$ variable reveals that every additional meter which the drone is placed away from the landing platform at the start, reduces the chances of making it to the landing platform by 21.01 percentage points. The  $\mathrm{TotalCuboidVolume}$ variable, on the other hand, is not statistically significant and we can thus conclude that the size of the obstacles didn't seem to impact whether drone made it to the landing platform.

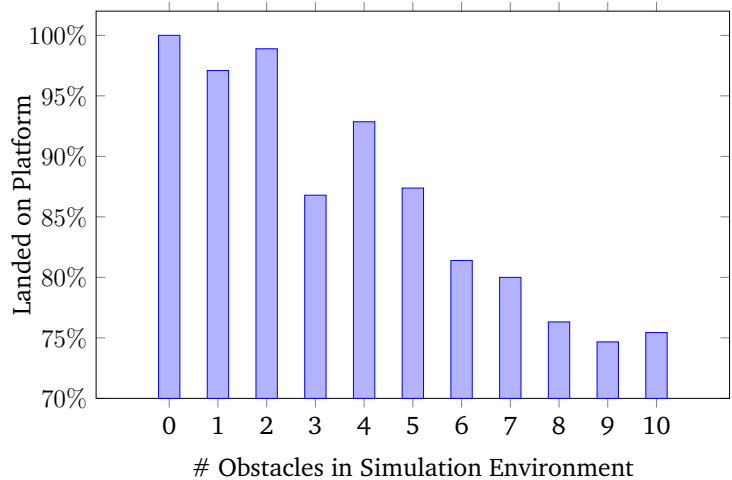
\begin{figure}[!htb]
\centering
\begin{tikzpicture}[scale=0.8]
\begin{axis}
[
width=12cm,
height=8cm,
xlabel={\# Obstacles in Simulation Environment}, 
xlabel style={yshift=-0.3cm},
xtick=data,
xticklabel style={rotate=0},
symbolic x coords={0, 1, 2, 3, 4, 5, 6, 7, 8, 9, 10},
ylabel={Landed on Platform},
ylabel style={yshift=0cm},
yticklabel={$\pgfmathprintnumber{\tick}\%$},
ymin=70, 
ymax=102, 
enlarge x limits=0.15,
ybar,
bar width=10,
]
\addplot coordinates {(0,100) (1,97.09) (2,98.89) (3,86.79) (4,92.86) (5,87.38) (6,81.4) (7,80.00) (8,76.32) (9,74.67) (10,75.44)};
\end{axis}
\end{tikzpicture}
\caption{Relative share of flights that landed on the platform depending on number of obstacles in the simulation environment}
\label{fig:hist_cuboids}
\end{figure}

\begin{figure}[!htb]
    \centering
    \begin{minipage}{.5\textwidth}
        \centering
        \includegraphics[width=0.9\linewidth]{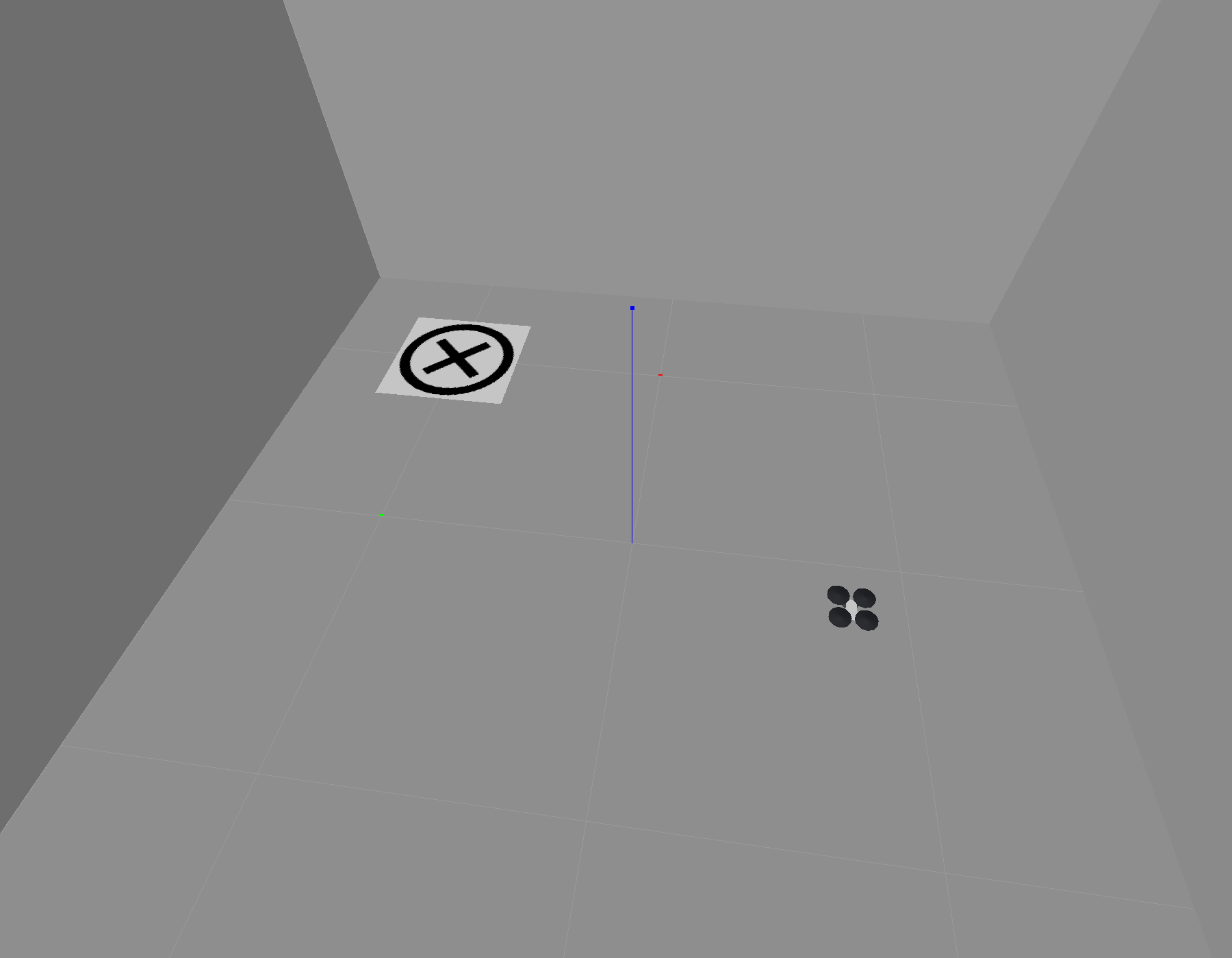}
        \caption{Example of non-obstructed environment}
        \label{fig:nonobs}
    \end{minipage}%
    \begin{minipage}{0.5\textwidth}
        \centering
        \includegraphics[width=0.9\linewidth]{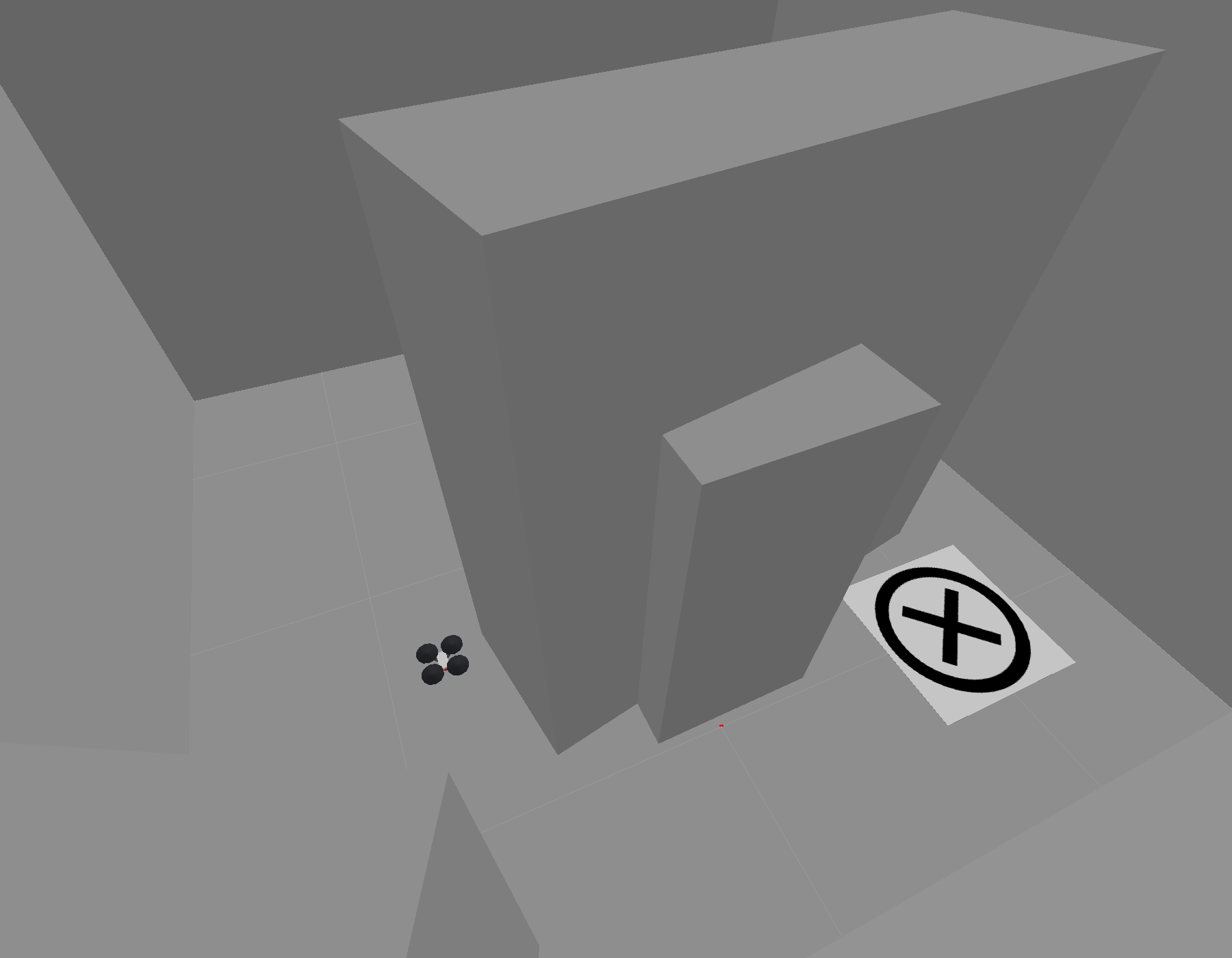}
        \caption{Example of obstructed environment (four obstacles)}
        \label{fig:obs}
    \end{minipage}
\end{figure}

Let's have a closer look at these two factors, beginning with the number of cuboids in the simulation environment. The bar chart depicted in Figure~\ref{fig:hist_cuboids} illustrates the share of test flights in which the model successfully managed to navigate the drone to the landing platform, depending on the number of obstacles in the simulation environment. There are two main takeaways from this figure: Firstly, we can see that in a simulation environment without any obstacles, the drone made it to the landing platform in every test flight. This corresponds to a success rate of 100\% and we can therefore conclude that the model achieves an optimal outcome in simulation environments without obstacles. Secondly, we can see that the chances of making it to the landing platform decreases with the number of obstacles, which confirms our results from the multi linear regression model. Figure~\ref{fig:hist_cuboids} illustrates that the model is 25\% less likely to navigate the drone to the landing platform in a highly obstructed simulation environment compared to a non-obstructed simulation environment. Examples of non-obstructed and obstructed simulation environments are given in Figure~\ref{fig:nonobs} and Figure~\ref{fig:obs}, respectively.

\begin{figure}[!htb]
\centering
\input{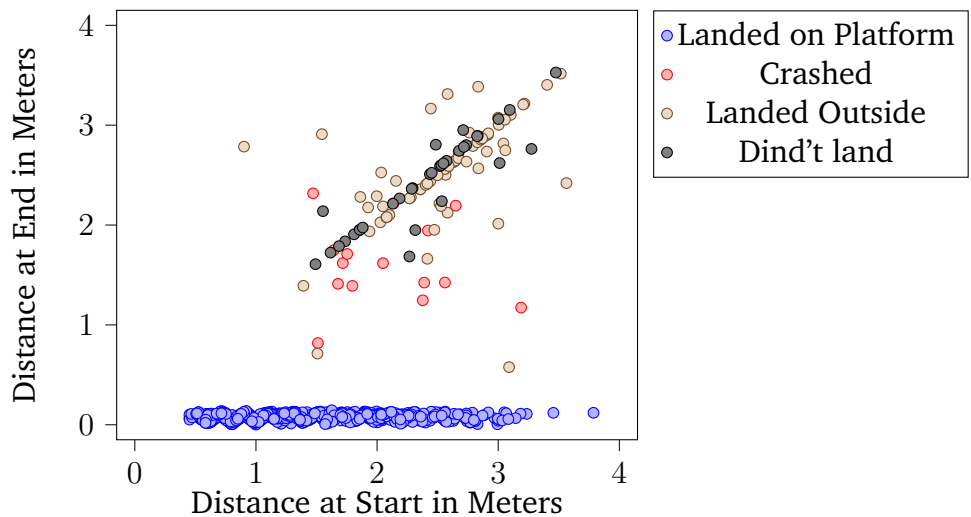}
\caption{Scatter plot comparing distance between position of the drone and position of the landing platform at the start and end of the test flights}
\label{fig:scatter_distance}
\end{figure}

The second factor, according to our multi linear regression model, which impacted the outcome of the test flights, was the distance between the starting position of the drone and the position of the landing platform. We have plotted the distance at the start against the distance at the end for each test flight in Figure~\ref{fig:scatter_distance}. The respective outcomes of the flights are color coded and reveal some interesting insights. First of all, we can see that whenever the drone is closer than 1.5 meters away from the landing platform at the start, it makes it to the landing platform in all but one of the test flights. Once the start distance is greater than 1.5 meters, the number of unsuccessful flights increases. Let's have a look at each of the four different outcome types. It is obvious that the final distance of flights in which the drone landed on the platform should be marginal. In fact, a flight is identified as ``Landed on Platfrom" when it lands within 30cm of the center of the landing platform. Looking at the flights which landed outside the landing platform, we can see that in most cases the drone landed far away from the platform. There are only three instances in which the drone landed outside the platform but within 1.5 meters. This finding, paired with the fact that the model achieved a 100\% success rate in the non-obstructed environments, suggests that whenever the landing platform is within the drone's visual field, it will successfully land on it. But whenever the drone needs to search for the landing platform, it can happen that it gets lost and lands somewhere else.  Regarding the test flights which didn't land at all, we can find that for most of these test flights, the distance at the end is equivalent to the distance at the start. We have observed that the drone sometimes takes off and begins rotating without ever leaving its position. Lastly, regarding the test flights which crashed, we can see that all but one of these flights crashed on their way towards the landing platform since the distance at the end is smaller than the distance at the start.

\subsection{Flight Path Analysis}

In order to evaluate the performance of the model, it is important to not only look at whether the drone made it to the platform, but also at the flight path which it has taken. We are interested in a model which manages to find the most efficient flight path and thus gets to the landing platform with as few flight command executions as possible. 

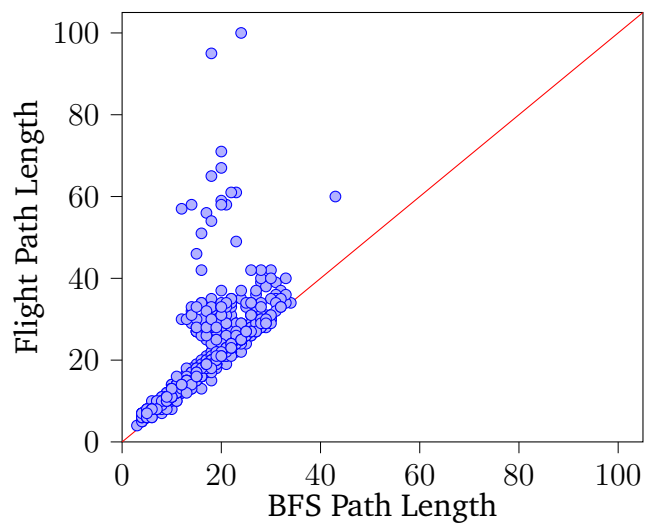
\begin{figure}[!htb]
\centering
\begin{tikzpicture}

\definecolor{color0}{rgb}{0.12156862745098,0.466666666666667,0.705882352941177}

\begin{axis}[
tick align=outside,
tick pos=left,
x grid style={white!69.0196078431373!black},
xlabel={BFS Path Length},
xmin=0, xmax=105,
xtick style={color=black},
y grid style={white!69.0196078431373!black},
ylabel={Flight Path Length},
ymin=0, ymax=105,
ytick style={color=black}
]
\draw [red] (0,0) -- (105,105);
\definecolor{b1}{RGB}{178, 178, 255}
\definecolor{b2}{RGB}{0, 0, 255}
\addplot [only marks, mark=*, draw=b2, fill=b1]
table{%
x                      y
14 15
18 19
28 29
28 28
17 19
22 23
12 14
24 35
14 17
10 14
19 19
21 22
17 56
16 18
20 26
30 30
12 30
19 31
15 17
12 12
7 9
18 33
25 27
26 29
5 6
17 19
19 21
21 58
34 34
13 15
22 22
21 19
22 24
18 22
13 15
12 13
29 30
27 28
22 24
21 23
18 19
11 11
16 18
16 34
6 8
24 26
7 9
15 16
14 15
20 21
15 16
8 9
12 13
12 15
10 13
17 29
7 9
13 15
26 28
25 26
12 13
26 27
25 27
29 28
10 13
9 11
16 17
10 11
18 34
4 5
10 13
13 13
11 13
13 14
20 59
6 9
10 11
5 6
20 21
24 25
10 12
21 23
22 34
30 30
25 26
20 22
4 5
18 19
12 15
26 27
25 28
29 30
18 27
13 14
30 42
5 7
6 8
16 17
19 22
21 21
11 14
9 11
21 25
15 16
20 37
15 16
24 28
14 14
17 18
10 12
11 13
10 12
15 17
8 10
18 19
8 9
12 14
19 20
18 65
23 22
19 19
24 22
18 20
19 22
10 11
26 27
14 16
21 24
11 13
5 8
4 7
20 22
20 23
6 7
18 19
12 14
17 19
6 8
14 13
4 5
20 67
17 18
16 34
11 13
15 16
10 14
16 13
16 42
10 11
19 25
6 8
23 27
18 20
10 11
24 25
19 20
13 14
5 6
8 11
12 14
12 13
8 8
26 27
14 29
9 11
12 14
9 10
13 15
15 16
19 20
22 23
18 35
26 33
19 21
7 8
25 27
13 15
26 34
6 7
23 25
17 21
16 18
13 15
8 9
10 11
25 26
22 24
22 30
19 19
5 7
6 8
12 15
8 9
21 24
14 15
10 12
15 16
19 25
21 30
20 24
28 39
22 22
9 11
18 19
9 12
17 19
27 36
19 26
6 8
18 24
15 19
12 14
31 39
18 19
28 28
5 6
30 31
10 11
26 28
32 37
23 61
6 8
14 16
17 28
21 22
16 18
17 18
19 20
14 15
17 18
28 29
9 10
24 28
12 57
10 8
27 32
20 22
14 15
26 28
28 29
11 13
12 14
5 7
29 30
6 8
19 24
24 26
13 14
8 10
7 9
5 7
11 12
13 14
15 16
6 7
11 13
15 16
19 20
11 12
25 24
10 12
17 19
7 8
20 27
18 19
11 13
9 11
18 19
21 23
29 38
14 16
4 6
27 28
7 8
16 20
23 24
18 20
20 22
17 18
24 100
20 23
15 16
18 19
16 18
20 58
25 26
6 8
12 14
22 23
15 16
16 17
30 31
18 21
6 8
6 8
10 11
16 17
23 49
11 12
4 7
23 24
16 17
10 12
12 13
12 12
21 22
15 17
17 17
19 21
5 7
17 20
5 7
23 25
15 16
13 15
18 54
10 12
15 27
9 11
6 8
22 23
8 10
12 13
26 31
31 36
22 24
28 30
20 22
7 8
26 31
19 19
25 28
5 6
5 7
17 19
19 21
10 11
8 10
11 12
24 25
18 19
27 37
23 25
22 22
18 19
11 13
30 30
18 19
26 26
11 13
18 19
16 17
14 16
26 27
12 14
20 22
19 21
13 14
24 26
16 18
20 21
16 17
17 18
15 16
25 26
10 12
29 30
21 22
10 12
15 15
18 30
17 29
19 26
10 13
10 12
15 17
20 22
15 27
11 12
12 14
17 19
19 20
13 15
20 25
20 30
9 10
24 25
12 14
11 14
17 19
20 21
20 21
19 21
20 28
24 27
10 13
22 23
22 23
19 27
29 31
20 22
13 14
16 17
18 28
25 26
21 24
17 18
24 26
17 19
10 12
12 14
22 24
20 31
15 46
24 26
14 17
18 20
12 14
28 40
22 31
21 22
11 12
22 33
23 29
25 27
28 30
20 22
22 28
19 21
12 14
16 18
18 19
27 28
9 11
6 7
11 14
14 14
17 18
16 17
19 18
13 15
18 15
10 12
26 27
13 15
18 22
6 6
25 28
15 16
16 18
5 6
16 16
18 33
19 30
19 20
14 31
11 14
15 17
16 18
22 24
4 7
21 22
6 10
17 18
13 14
17 19
22 23
16 30
15 27
23 26
11 13
21 21
24 25
25 27
15 17
28 28
19 19
17 25
17 19
13 30
29 30
10 12
30 31
22 23
27 33
13 17
17 19
18 20
3 4
24 25
13 14
19 21
19 27
19 22
6 8
15 17
16 16
9 10
14 15
16 17
18 20
12 13
10 10
33 34
16 17
21 23
14 32
24 25
7 9
15 28
15 16
26 26
14 18
15 16
17 17
16 17
6 8
19 25
30 30
22 35
9 12
21 23
12 14
23 24
22 23
16 17
21 31
28 42
4 5
5 7
22 28
15 18
11 13
16 26
12 15
11 12
10 13
31 35
21 25
12 12
12 14
33 36
19 21
22 34
9 11
26 31
21 23
21 26
13 14
8 10
24 28
7 9
9 12
20 22
11 12
30 29
9 11
13 18
11 12
6 8
7 10
10 11
30 35
33 40
16 17
19 21
14 15
18 20
9 10
18 28
4 6
20 33
4 6
24 26
18 18
23 25
15 30
30 40
18 20
17 18
21 22
14 17
15 17
21 23
6 8
10 12
27 27
21 22
14 33
5 7
13 15
11 12
28 31
26 29
22 23
8 10
27 27
4 6
19 21
8 10
11 13
24 28
24 29
29 29
9 11
8 11
13 14
32 35
25 26
10 12
17 27
11 10
14 14
18 19
9 10
13 15
15 17
19 21
21 23
16 19
20 24
26 28
17 20
13 15
5 6
18 17
19 22
18 20
17 32
18 19
26 27
16 16
11 10
15 16
27 30
14 13
26 42
21 22
18 19
31 34
27 29
28 34
5 7
5 6
10 12
13 15
21 22
29 30
9 11
9 10
19 19
11 13
8 10
9 11
12 14
29 30
12 13
17 19
9 9
18 21
17 28
24 28
18 20
8 9
25 27
4 7
9 11
25 26
12 14
24 37
5 8
24 25
12 15
5 8
20 33
7 9
24 24
15 17
11 12
15 18
18 19
14 15
21 22
18 19
14 16
8 10
13 14
10 12
22 23
11 14
14 31
22 23
9 11
27 28
8 10
10 11
18 19
22 23
24 25
14 58
7 10
10 11
14 15
16 17
16 18
8 7
23 24
28 29
20 71
15 17
15 17
25 27
25 27
10 13
16 17
31 32
13 14
19 20
18 20
12 14
14 15
15 16
18 20
22 21
20 31
6 8
23 25
16 17
8 8
21 29
7 8
9 10
21 23
16 18
11 12
22 61
26 27
18 29
16 17
18 24
9 8
15 16
13 14
14 17
13 15
16 18
12 14
23 29
13 14
24 25
26 28
5 6
19 22
6 8
19 29
11 14
15 17
22 23
8 9
15 15
20 33
18 19
16 16
30 31
13 14
18 19
25 33
13 16
21 33
19 25
26 34
5 6
27 29
5 7
15 16
13 15
26 27
16 51
22 23
10 11
15 28
24 25
21 23
5 6
19 20
15 33
28 30
11 12
17 26
9 11
9 11
17 18
15 16
15 18
21 29
22 23
16 18
43 60
32 33
18 20
15 16
27 27
13 15
28 30
22 24
6 6
13 14
18 19
22 35
20 34
21 22
8 10
6 8
15 17
17 19
17 28
28 29
25 34
21 34
18 30
24 26
19 28
21 23
29 30
13 15
19 20
32 33
13 15
9 11
20 22
24 29
11 12
10 13
17 19
12 13
29 29
20 31
23 27
23 25
23 25
9 10
13 13
26 27
22 23
9 10
14 14
7 8
23 26
25 27
22 24
22 24
26 31
26 34
15 17
20 21
18 95
9 10
10 11
6 8
10 11
22 23
24 25
15 16
19 21
11 16
10 13
9 11
24 25
20 21
20 33
13 12
25 27
18 18
12 14
28 33
17 19
5 7
12 14
};
\end{axis}

\end{tikzpicture}
\caption{Scatter plot comparing BFS path length with flight path length for test flights which landed on the platform}
\label{fig:flight_path}
\end{figure}

For a first analysis, we only considered the test flights in which the drone landed on the platform. Figure~\ref{fig:flight_path} includes a scatter plot in which the flight path length is plotted against the path length that has been determined by our BFS algorithm described in Section~\ref{sec:bfs}. We can see that the majority of flight paths is close to the BFS flight paths. In fact, 71\% of the flight paths have a length which is less than 20\% longer than the corresponding BFS path. We can even see that the flight paths taken by the model are in some cases marginally shorter than the BFS paths. This can happen because we are using some simplifications when calculating the BFS path. Figure~\ref{fig:flight_path} also illustrates that the variance of the flight path length increases with the length of the corresponding BFS path. Longer BFS paths suggest that the drone might not be able to see the landing from the beginning and first needs to manoeuvre around some obstacles in order to find it. In these cases, the drone sometimes takes detours which explains the increasing variance. 

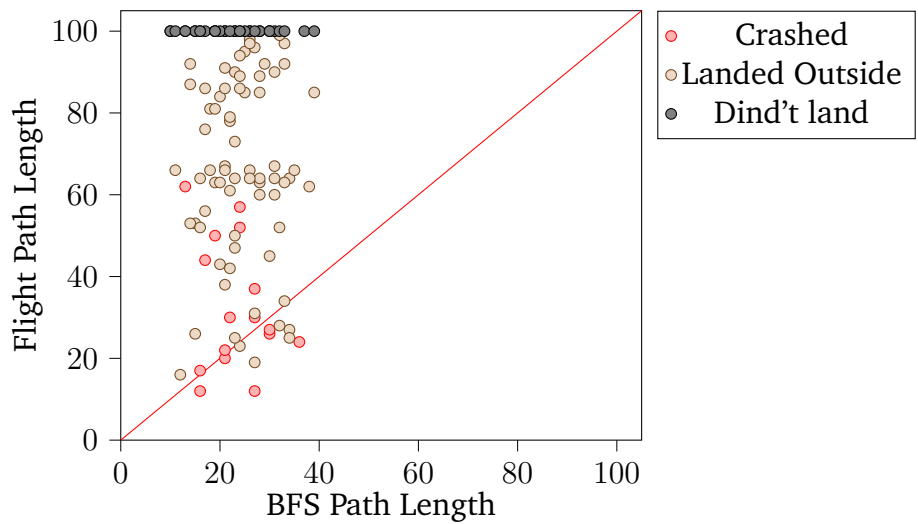
\begin{figure}[!htb]
\centering
\begin{tikzpicture}

\definecolor{color0}{rgb}{0.12156862745098,0.466666666666667,0.705882352941177}

\begin{axis}[
tick align=outside,
tick pos=left,
x grid style={white!69.0196078431373!black},
xlabel={BFS Path Length},
xmin=0, xmax=105,
xtick style={color=black},
y grid style={white!69.0196078431373!black},
ylabel={Flight Path Length},
ymin=0, ymax=105,
ytick style={color=black},
legend pos=outer north east
]
\draw [red] (0,0) -- (105,105);
\definecolor{r1}{RGB}{255, 178, 178}
\definecolor{r2}{RGB}{255, 0, 0}
\addplot [only marks, mark=*, draw=r2, fill=r1]
table{%
x                      y
19 50
21 20
16 12
24 52
17 44
24 57
21 22
36 24
27 30
27 12
13 62
30 26
22 30
27 37
30 27
16 17
};
\definecolor{g1}{RGB}{236, 217, 198}
\definecolor{g2}{RGB}{115, 77, 38}
\addplot [only marks, mark=*, draw=g2, fill=g1]
table{%
x                      y
29 92
18 81
17 86
39 85
23 90
33 92
31 64
26 66
21 67
31 90
20 43
34 64
25 85
23 64
16 64
15 53
14 53
21 91
17 76
19 81
34 27
33 97
31 67
24 89
32 52
23 47
17 56
24 23
28 63
21 38
28 89
15 26
35 66
19 63
12 16
28 64
21 66
27 31
26 64
30 100
25 95
38 62
30 45
16 52
27 96
24 86
23 73
11 66
14 92
23 50
20 63
26 98
32 99
20 84
34 25
26 97
33 63
31 60
33 34
22 78
28 60
32 28
18 66
21 86
23 25
27 19
22 79
14 87
22 61
24 94
22 42
28 85
};
\definecolor{z1}{RGB}{128, 128, 128}
\definecolor{z2}{RGB}{0, 0, 0}
\addplot [only marks, mark=*, draw=z2, fill=z1]
table{%
x                      y
30 100
28 100
10 100
21 100
10 100
21 100
23 100
17 100
26 100
37 100
15 100
16 100
26 100
28 100
13 100
31 100
27 100
23 100
19 100
39 100
20 100
30 100
32 100
25 100
24 100
19 100
11 100
15 100
13 100
19 100
22 100
16 100
33 100
24 100
};
\legend{Crashed, Landed Outside, Dind't land}
\end{axis}

\end{tikzpicture}
\caption{Scatter plot comparing BFS path length with flight path length for test flights which did not landed on the platform}
\label{fig:flight_path_not_landed}
\end{figure}

Using the scatter plot depicted in Figure~\ref{fig:flight_path_not_landed}, we can conduct the same analysis for all test flights in which the drone didn't land on the landing platform. Let's again go through each of the three different outcomes separately, beginning with the test flights in which the drone crashed. We can see that the crashes didn't occur right at the start, but rather after at least twelve flight command executions. In four of these flights, the drone even crashed after having executed over 50 flight commands already. Regarding the flights in which the drone landed outside the landing platform, we can see that in the majority of these test flights, the drone landed after having executed more than 60 flight commands. Since the model takes in the number of previously executed flight commands as an input, it makes sense that it becomes more likely for the model to predict the landing command after having executed more flight commands even if the landing platform isn't yet in sight. Nevertheless, there were also test flights in which the drone didn't land after 100 flight command executions. In regards to the flights which didn't land at all, Figure~\ref{fig:flight_path_not_landed} doesn't reveal any insights except for the fact that  this outcome is evenly distributed across the BFS path lengths.

\section{Qualitative Comparison}

After having evaluated the performance of our model, we will now qualitatively compare our approach with the Reinforcement Learning approach of \citet{polvara} and the Supervised Learning approaches of \citet{kim} and \citet{ram}. 

All three papers \citep{kim, ram, polvara} attempt to solve a task which is similar to ours (i.e. vision-based autonomous navigation and landing of a drone) but differs in regards to some aspects (e.g. \citet{kim} and \citet{ram} don't use a landing platform and also don't include obstacles in the environment; \citet{polvara} use a downward facing camera and fly the drone at higher altitudes). 

In regards to how these papers \citep{kim, ram, polvara} approach to solve this task, we can find some similarities but also some key differences. What our approach and all three papers \citep{kim, ram, polvara} have in common, is the fact that we all use a CNN to map images that are captured by the drone to high-level control commands. Let's discuss the other similarities and differences separately for the Reinforcement Learning approach of \citet{polvara} and the the Supervised Learning approaches of \citet{kim} and \citet{ram}. The key points of our comparison are summarized in Table~\ref{tab:comp}.

\begin{table}[!htb]
\centering
\caption{Qualitative comparison of our approach with previous literature}
\begin{threeparttable}
\begin{tabular}{m{3cm} | m{3cm} | m{3cm} | m{3cm}}
\hline
\hline
 & \textbf{\citep{polvara}} & \textbf{\citep{kim} and \citep{ram}} & \textbf{Our approach} \\
\hline
\textbf{Learning} & Reinforcement & Supervised & Supervised \\
\hline
\textbf{Training \mbox{Environment}} & Simulation & Real World & Simulation \\
\hline
\textbf{Data Collection} & - & Manual & Automated \\
\hline
\textbf{Convergence} & slow & quick & quick \\
\hline
\textbf{Generalisation} & Domain \mbox{Randomisation} & Augmentation & (Domain \mbox{Randomisation})\tnote{1} \\
\hline
\hline
\end{tabular}
\begin{tablenotes}
\item[1] Domain randomisation has not been implemented yet, but our approach can be extended to use domain randomisation for sim-to-real transfer.
\end{tablenotes}
\end{threeparttable}
\label{tab:comp}
\end{table}

What the approach of \citet{polvara} has in common with our approach, is the fact that we both use a simulation program for training the machine learning model. Our approaches do however differ in regards to \textit{how} we use the simulation program. We use the simulation program to collect labelled data samples and subsequently train our model via Supervised Learning. In contrast, \citet{polvara} apply an enhance reward method (partitioned buffer replay) for training their machine learning model via Reinforcement Learning in the simulator. A potential downside of their approach is the low convergence rate. \citet{polvara} were in fact required to train their model for 12.8 days in simulation. Using our approach in contrast, it takes 15.6 hours to generate a dataset of 1\,000\,000 data samples using an Intel Core i7-7700K CPU with 16 GB random access memory (RAM) and 1.2 hours to train the model using a NVIDIA GeForce GTX 1080 GPU with 3268 MB RAM. The significantly quicker convergence rate constitutes an advantage of our approach compared to the approach of \citet{polvara}. An aspect in which the approach of \citet{polvara} is more advanced than ours, concerns sim-to-real transfer. \citet{polvara} trained their model in domain randomised simulation environments with varying visual and physical properties. This allowed them to successfully transfer their learnings into the real world \citep{polvara}. We have not implemented domain randomisation features yet, but our approach can be extended with domain randomisation features to allow for sim-to-real transfer.

\citet{kim} and \citet{ram} use a Supervised Learning technique for training their CNN. Their convergence rate is therefore similar to ours. The main difference compared to our approach concerns the data collection procedure that is used for generating a labelled dataset. Both, \citet{kim} and \citet{ram} manually collect images in the real world. This is a time consuming task as they need to fly the real drone through various locations and label them with the corresponding flight command that is executed by the pilot \citep{kim, ram}. In order to increase the size of their datasets, \citet{kim} and \citet{ram} further apply augmentation techniques such as zooming, flipping and adding Gaussian white noise. We, in contrast, have developed a fully automated procedure to collect data samples. In our approach, images and sensor information are collected in a simulation program and the labels, which correspond to the optimal flight commands, are computed using a path planning algorithm. Under the assumption that sim-to-real transfer is possible, which has been shown by several other publications \citep{tobin, johns, sadeghi, kaufmann, polvara}, our approach has two advantages: Firstly, since the data collection procedure is fully automated, we can create large datasets with very little effort and thus generalize the model more by training it on a greater variety of environments. Secondly, our labelling is more accurate since we don't have humans in the loop and our path planning algorithm ensures that the label corresponds to the optimal flight command.

\section{Summary}

This chapter gave an overview of the test results that were achieved in the simulation environment, and provided a qualitative comparison to previous publications.

The test results revealed that the model managed to successfully navigate the drone towards the landing platform in 87.8\% of the test flights. We have shown that the risk of crashing the drone can be reduced when training the model on larger datasets, even though we couldn't find a consistent improvement in regards to successful landings. Since each test flight was performed in a different simulation environment, it was also possible to observe the impact of environmental changes on performance of the model. The results illustrated that the success rate negatively correlates with the number obstacles in the environment as well as the distance between the starting position of the drone and the position of the landing platform. The model achieved a success rate of 100\% in non-obstructed simulation environments. 

In regards to the flight path, we have found that the flight path taken by the model is for the majority of test flights very close to the optimal flight path. It can, however, sometimes happen that the drone takes a detour especially when the simulation environment is highly obstructed.

The qualitative comparison of our model with previous research has shown that our model has the advantage of having shorter training times when being compared to a similar Reinforcement Learning approach and overcomes the limitations of manual data collection (under the assumption of successfully applying domain randomisation for sim-to-real transfer) when being compared to similar Supervised Learning approaches.

\chapter{Conclusion}

In this work, we have presented a vision-based control approach which maps low resolution image and sensor input to high-level control commands. The task which this model attempts to solve is the autonomous navigation of a MAV towards a visually marked landing platform in an obstructed indoor environment. Our approach is practical since (i) our data collection procedure can be used to automatically generate large and accurately labelled datasets in simulation and (ii) because our model has a quick convergence rate. We expect our approach to also be applicable to other navigation tasks such as the last meter autonomous control of a delivery robot or the navigation of underwater vehicles. Despite the simplicity of our approach, the test flights in the simulation environment showed very promising results. There are several possibilities for future work:

\begin{itemize}
\item{Extend the approach by applying domain randomisation techniques (e.g. the ones used by \citet{johns}) to explore whether it is possible to transfer the learnings into the real world.}
\item{Make the approach suitable for dynamic environments (e.g. extend the simulation program to also include dynamic features; adapt the path planning algorithm to account for changes in the environment).}
\item{Enhance the approach to make continuous control decisions based on a video stream (e.g. using DJI Tello's set commands).}
\item{Use hardware acceleration (e.g. FPGAs) to speed up the drone simulator.}
\end{itemize}

 \bibliographystyle{unsrtnat}
\bibliography{bibliography}


\begin{appendices}
\chapter{Extrinsic Camera Matrix}
\label{app:extr}

\begin{minipage}{\linewidth}
\begin{lstlisting}[caption={Function for computation of extrinsic camera matrix}, label={lst:extrinsic_matrix}, language=Python, style=python]
    def calculate_view_matrix(self):
        """
        Calculates the extrinsic matrix of a camera during simulation.

        Parameters
        ----------
        no paramters

        Returns
        -------
        view_matrix: list of 16 floats
            The 4x4 view matrix corresponding to the position and orientation of
            the camera
        """
        # get current position and orientation of link
        camera_position, camera_orientation, _, _, _, _ = p.getLinkState(
            self.body_id, self.link_id, computeForwardKinematics=True)
        # calculate rotational matrix
        rotational_matrix = p.getMatrixFromQuaternion(
            camera_orientation)
        rotational_matrix = np.array(
            rotational_matrix).reshape(3, 3)
        # transform camera vector into world coordinate system
        camera_vector = rotational_matrix.dot(
            self.initial_camera_vector)
        # transform up vector into world coordinate system
        up_vector = rotational_matrix.dot(
            self.initial_up_vector)
        # compute view matrix
        view_matrix = p.computeViewMatrix(
            camera_position, camera_position + camera_vector, up_vector)
        return view_matrix
\end{lstlisting}
\end{minipage}

\chapter{Datasets used for Optimisation}
\label{app:dataset}

\begin{figure}[!htb]
\centering
\begin{tikzpicture}[scale=0.8]
\begin{axis}
[
xlabel={Label}, 
xlabel style={yshift=-0.3cm},
xtick=data,
xticklabel style={rotate=0},
symbolic x coords={takeoff, land, forward, cw, ccw},
ylabel={\#Samples},
ylabel style={yshift=0.5cm},
ymin=0, 
ymax=32000, 
enlarge x limits=0.15,
nodes near coords, 
nodes near coords align={vertical}, 
ybar,  
]
\addplot coordinates {(takeoff,3608) (land,3524) (forward,27220) (cw,18129) (ccw,17509) };
\end{axis}
\end{tikzpicture}
\caption{Histogram of train dataset}
\label{fig:hist_train}
\end{figure}
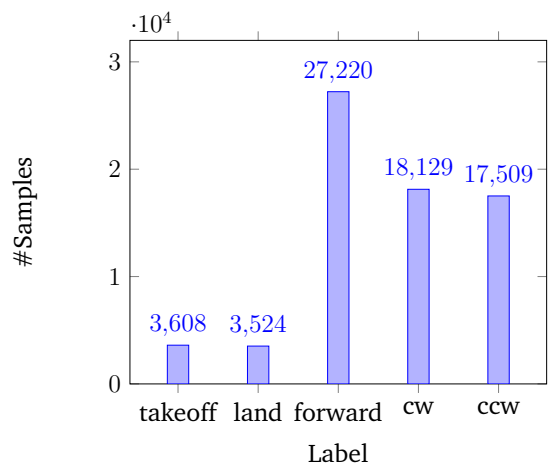

\begin{figure}[!htb]
\centering
\begin{tikzpicture}[scale=0.8]
\begin{axis}
[
xlabel={Label}, 
xlabel style={yshift=-0.3cm},
xtick=data,
xticklabel style={rotate=0},
symbolic x coords={takeoff, land, forward, cw, ccw},
ylabel={\#Samples},
ylabel style={yshift=0.5cm},
ymin=0, 
ymax=7000, 
enlarge x limits=0.15,
nodes near coords, 
nodes near coords align={vertical}, 
ybar,  
]
\addplot coordinates {(takeoff,749) (land,733) (forward,5884) (cw,3862) (ccw,3772) };
\end{axis}
\end{tikzpicture}
\caption{Histogram of validation dataset}
\label{fig:hist_val}
\end{figure}
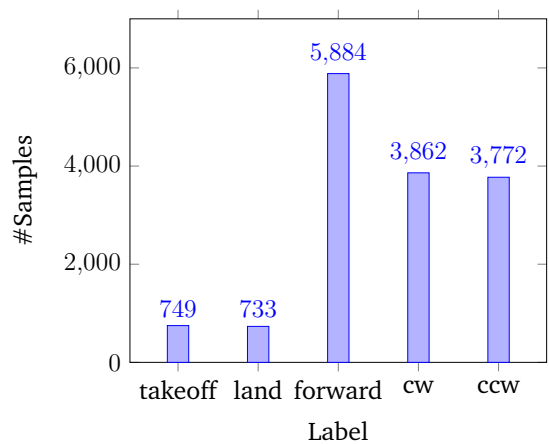

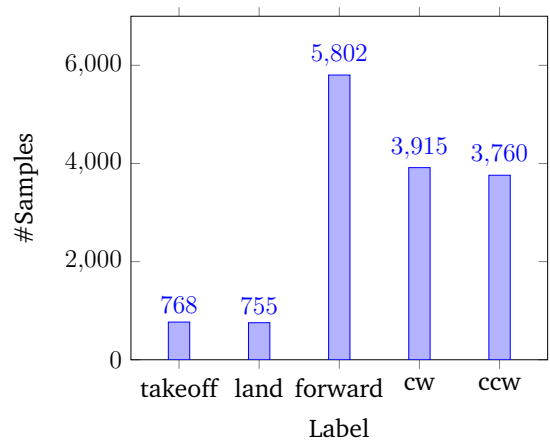
\begin{figure}[!htb]
\centering
\begin{tikzpicture}[scale=0.8]
\begin{axis}
[
xlabel={Label}, 
xlabel style={yshift=-0.3cm},
xtick=data,
xticklabel style={rotate=0},
symbolic x coords={takeoff, land, forward, cw, ccw},
ylabel={\#Samples},
ylabel style={yshift=0.5cm},
ymin=0, 
ymax=7000, 
enlarge x limits=0.15,
nodes near coords, 
nodes near coords align={vertical}, 
ybar,  
]
\addplot coordinates {(takeoff,768) (land,755) (forward,5802) (cw,3915) (ccw,3760) };
\end{axis}
\end{tikzpicture}
\caption{Histogram of test dataset}
\label{fig:hist_test}
\end{figure}

\chapter{Architecture Comparison}
\label{app:arch_comp}

\begin{table}[!htb]
\centering
\caption{Parameter values used for architecture comparison}
\begin{tabular}{c c}
\hline
\hline
\textbf{Parameter} & \textbf{Value} \\
\hline

\# Epochs & 10 \\
Batch Size & 32 \\
\\
\# Conv. Filters & 32 \\
Conv. Kernel Size & 3 x 3 \\
Conv. Padding & valid \\
Conv. Stride & 1 \\
Conv. Activation Function & ReLU \\
\\
Pool. Kernel Size & 2 x 2 \\
Pool. Padding & valid \\
Pool. Stride & 1 \\
\\
\# Neurons & 32 \\
Fully Con. Layer Activation Function & ReLU \\

\hline
\hline
\end{tabular}
\label{tab:intrinsic_parameters}
\end{table}

\chapter{Multi Linear Regression Output}
\label{app:regression}

\begin{table}[!htb]
\caption{Multi linear regression output}
\resizebox{\linewidth}{!}{%
\begin{tabular}{lclc}
\toprule
\textbf{Dep. Variable:}        & landed\_on\_platform & \textbf{  R-squared:         } &     0.290   \\
\textbf{Model:}                &         OLS          & \textbf{  Adj. R-squared:    } &     0.288   \\
\textbf{Method:}               &    Least Squares     & \textbf{  F-statistic:       } &     135.6   \\
\textbf{Date:}                 &   Wed, 26 Aug 2020   & \textbf{  Prob (F-statistic):} &  1.15e-73   \\
\textbf{Time:}                 &       08:43:40       & \textbf{  Log-Likelihood:    } &   -130.77   \\
\textbf{No. Observations:}     &          1000        & \textbf{  AIC:               } &     269.5   \\
\textbf{Df Residuals:}         &           996        & \textbf{  BIC:               } &     289.2   \\
\textbf{Df Model:}             &             3        & \textbf{                     } &             \\
\bottomrule
\end{tabular}}
\\
\resizebox{\linewidth}{!}{%
\begin{tabular}{lcccccc}
                               & \textbf{coef} & \textbf{std err} & \textbf{t} & \textbf{P$> |$t$|$} & \textbf{[0.025} & \textbf{0.975]}  \\
\midrule
\textbf{Intercept}             &       1.3329  &        0.024     &    54.553  &         0.000        &        1.285    &        1.381     \\
\textbf{nr\_cuboids}           &      -0.0315  &        0.004     &    -7.040  &         0.000        &       -0.040    &       -0.023     \\
\textbf{total\_cuboid\_volume} &       0.0023  &        0.004     &     0.607  &         0.544        &       -0.005    &        0.010     \\
\textbf{distance\_start}       &      -0.2101  &        0.012     &   -17.486  &         0.000        &       -0.234    &       -0.187     \\
\bottomrule
\end{tabular}}
\end{table}

\end{appendices}

\end{document}